\documentclass[APA,STIX2COL]{WileyNJD-v2}
\usepackage{moreverb}
\usepackage{xcolor}
\usepackage{adjustbox}
\usepackage{subcaption}
\usepackage{mathtools}
\usepackage{balance}
\usepackage{amsmath,amsfonts}
\usepackage{bm}
\usepackage{multirow}
\usepackage{amssymb}
\usepackage{enumitem}
\usepackage{caption}
\usepackage{array}
\usepackage[doipre={~}]{uri}
\usepackage{xcolor, soul}
\sethlcolor{yellow}

\usepackage{mathtools}
\usepackage{nomencl}

\makenomenclature

\renewcommand\figurename{\textbf{FIGURE}}
\renewcommand{\thefigure}{\arabic{figure}}

\makeatletter
\def\NAT@aysep{,}
\makeatother

\articletype{Original Article}%

\received{26 February 2021}
\revised{10 May 2021}
\accepted{10 June 2021}

\begin{document}

\title{Balanced Semi-Supervised Generative Adversarial Network for Damage Assessment from Low-Data Imbalanced-Class Regime}

\author[1,2]{Yuqing Gao}

\author[1,3]{Pengyuan Zhai}

\author[1,2,4]{Khalid M. Mosalam*}

\authormark{\textsc{Gao et al.}}

\address[1]{\orgdiv{Department of Civil and Environmental Engineering}, \orgname{University of California, Berkeley}, \orgaddress{\state{CA}, \country{USA}}}

\address[2]{\orgdiv{Tsinghua-Berkeley Shenzhen Institute (TBSI)}, \orgname{Tsinghua University}, \orgaddress{\state{Shenzhen}, \country{China}}}

\address[3]{\orgdiv{Department of Industrial Engineering and Operations Research}, \orgname{University of California, Berkeley}, \orgaddress{\state{CA}, \country{USA}}}

\address[4]{\orgdiv{Pacific Earthquake Engineering Research Center}, \orgname{University of California, Berkeley}, \orgaddress{\state{CA}, \country{USA}}}

\corres{Khalid M. Mosalam, 723 Davis Hall, University of California, Berkeley, CA 94720-1710, USA. \email{mosalam@berkeley.edu}}

\abstract[Summary]{In recent years, applying deep learning (DL) to assess structural damages has gained growing popularity in vision-based structural health monitoring (SHM). However, both data deficiency and class-imbalance hinder the wide adoption of DL in practical applications of SHM. Common mitigation strategies include transfer learning, over-sampling, and under-sampling, yet these ad-hoc methods only provide limited performance boost that varies from one case to another. In this work, we introduce one variant of the Generative Adversarial Network (GAN), named the balanced semi-supervised GAN (BSS-GAN). It adopts the semi-supervised learning concept and applies balanced-batch sampling in training to resolve low-data and imbalanced-class problems. A series of computer experiments on concrete cracking and spalling classification were conducted under the low-data imbalanced-class regime with limited computing power. The results show that the BSS-GAN is able to achieve better damage detection in terms of recall and $F_\beta$ score than other conventional methods, indicating its state-of-the-art performance.}


\jnlcitation{\cname{%
\author{Gao Y.},
\author{Zhai P.}, and
\author{Mosalam K.M.}}
\ctitle{Balanced Semi-Supervised GAN in Structural Damage Assessment from Low-Data Imbalanced-Class Regime}. \cjournal{Computer-aided Civil and Infrastructure Engineering}, \cvol{2020; 00:1--14}.}

\maketitle

\section{Background \& Motivations}
Nowadays, machine learning (ML) and deep learning (DL) have been at the forefront of a plethora of research activities in many disciplines. Since 2017, there has been an increasing number of DL studies and applications in civil and structural engineering \citep{Rafiei2017, oh2017evolutionary, gao2018transfer,yeum2018visual, mosalam2019new, perez2019recurrent, amezquita2018wireless}. Particularly, for vision-based structural health monitoring (SHM), the convolutional neural network (CNN) is proven to be a promising approach with high practical potential \citep{cha2017deep, deng2020concrete, gao2018transfer, gao2018residual, liang2019image, maeda2016lightweight,jiang2019real, zhang2017automated, zhang2020concrete}. However, the frameworks in past studies are facing three real-world challenges: (1) insufficient data, (2) imbalanced classes, and (3) high computing demands when computing power is limited.

Most of the current DL applications in vision-based SHM fall into the supervised learning category, e.g., image classification \citep{gao2018transfer}, damage localization \citep{cha2018autonomous, xue2018fast} and segmentation \citep{yang2018automatic}, which requires substantial amounts of labeled data to reach the desired performance level. Obtaining large-scale labeled structural image datasets is costly and labor-intensive. Compared with popular computer vision (CV) benchmark datasets such as ImageNet (14 million images) \citep{imagenet09}, MNIST (70,000 images) \citep{lecun1998gradient} and CIFAR-10 (70,000 images) \citep{krizhevsky2009learning}, current vision-based SHM datasets, e.g., PEER Hub ImageNet ($\phi$-Net) with 36,413 images \citep{gao2020peer}, are much smaller. Although many previous studies \citep{cha2017deep, dorafshan2018comparison} heavily relied on cropping techniques to augment the datasets, the cropped images had poor variety of invariant features, as they were sourced from similar scenarios with limited numbers of raw images. Additionally, high similarities between cropped images in the training and test sets pose the risk of data leakage, as the model will memorize features in the training set that are simply repeated in the test set, which may exaggerate the model performance in real-world applications. It is to be noted that, herein, the term ``insufficient'' has a dual meaning: the lack of data quantity and the lack of feature variety.

There also exists the imbalanced-class issue, which stems from the very nature of SHM where structural damages (due to either natural deterioration or extreme events such as earthquakes) are low-frequency occurrences. In real-world SHM image collection processes, damage-related data (e.g., cracking and spalling) usually make up a small portion, and thus the majority of data belong to the undamaged state. This leads to imbalanced datasets and further causes the model to be biased in favor of the ``undamaged'' class. On top of the already-existing low-data issue, the small numbers of ``damaged'' data easily lead to model overfitting (especially for high-dimensional feature space of image data). The impact of imbalanced datasets on DL performance in SHM has not been thoroughly studied, as many previous studies avoided this issue by constructing and training on balanced datasets, which are small in size as previously mentioned.

Besides the two major issues discussed in previous paragraphs, computing power limitations create another daunting challenge for efficiently training classification models. DL has benefited from the advancement of high-performance Graphics Processing Unit (GPU), the lack of which however becomes an insuperable obstacle in real-world applications. For example, the limited payload/carry-on capacity of a given small unmanned aerial vehicle (UAV) \citep{villa2016overview} due to budget limits or external environmental factors forbids the deployment of high-performance GPUs and supporting modules. Thus, to be able to conduct real-time recognition/inference with limited hardware, the network architecture needs to be degraded to a shallow, lightweight design, e.g., MobileNet \citep{howard2017mobilenets}, which, however, will compromise the classification performance. Besides, it is also worth noting that to pursue a better recognition performance, the network architectures (e.g., number of layers and number of filters) in past studies were usually designed and tuned specifically for their respective datasets after many trials and iterations, but tuning model parameters is often costly, and may not be generic enough for practical applications.

Transfer learning (TL), over-sampling by conventional data augmentation (DA) methods, under-sampling, and weighted loss functions are common strategies to address the low-data and imbalanced-class issues. In TL, by tuning parameters from a pre-trained model, the new model can adapt to the target domain relatively easily, where the parameters in the early layers inherit certain knowledge from basic features in the source domain, making the model less dependent on extensive amounts of data \citep{pan2009survey}. However, TL requires a pre-trained and open-source model from a source dataset, e.g., VGGNet \citep{simonyan2014very} and ResNet  \citep{he2016deep} trained from the ImageNet dataset \citep{imagenet09}. For customized (self-designed) networks, pre-trained networks are sometimes inaccessible and usually computationally expensive since they have to be pre-trained on large-scale datasets, e.g., ImageNet \citep{imagenet09}, containing millions of training data. Moreover, TL only aims to mitigate the data deficiency problem and may not be sufficient to address the imbalanced-class issues \citep{weiss2017comparing}.

In over-sampling, the minority-class data are over-sampled to reduce the majority-to-minority ratio by randomly duplicating minority-class samples or performing certain transformations, i.e., pre-processing operations, e.g., translation, flip, scale, whitening, and adding noise. The over-sampled minority class data are then mixed with the majority-class data to build a relatively balanced dataset \citep{he2009learning}. However, conventional DA can only generate highly-correlated data, which does not increase feature variety, and additional storage space is needed for the over-sampled data. For some cases, inappropriately settings of DA might even lead to performance drops \citep{gao2020peer}. 

In under-sampling, majority-class data are randomly dropped to reduce the majority-to-minority ratio, which forms a balanced dataset with smaller size \citep{he2009learning}. However, under-sampling may lead to an untrainable situation of the DL model or performance degradation due to information loss by ruling out large amounts of data.

Unlike the aforementioned methods to manipulate either model parameters or data, the weighted loss function method is more straightforward. It aims to penalize misclassification of minority classes by assigning them higher class weights, while reducing the weights of majority classes. Such mechanism takes into account the skewed distribution of the classes. In practice, especially in object detection and semantic segmentation, balanced cross entropy and focal loss are widely adopted \citep{lin2017focal, cui2019class, eigen2015predicting}.

Besides the conventional methods discussed above, Generative Adversarial Network (GAN) is considered as an alternative DA method, where GAN generates synthetic data to augment the existing dataset and potentially enhance the model's performance \citep{goodfellow2014generative, salimans2016improved, madani2018semi}. Compared to applying TL to fine-tune a heavily-parameterized pre-trained DL models from ImageNet, a common GAN has simpler architecture and fewer trainable parameters, which makes GANs more applicable to custom-designed lightweight networks and less demanding for computing power. Compared with conventional DA, GANs can generate new data unseen by the model, and increase data variety accordingly.

In addition, some recent studies revealed the effectiveness of GAN in relieving the imbalanced-class issue. \cite{antoniou2017data} proposed data augmentation GAN (DAGAN), which overcomes the target domain's class imbalance issue by finding representations of the source domain meaningful to generate other related data and augmenting the target domain by new samples generated from the learnt manifold, which achieved promising enhancements. \cite{mariani2018bagan} proposed balancing GAN (BAGAN) as an augmentation method to restore balance in imbalanced datasets such that it can learn useful features from the majority classes and use these to generate images for the minority classes. BAGAN was shown to be more superior than ordinary GANs in terms of generated minority-class image quality and variability with imbalanced datasets. \cite{zhang2019dada} introduced deep adversarial data augmentation (DADA) and formulated the DA problem into a supervised class-conditional GAN by developing a $2K$ discriminator loss function where $K$ is the number of real classes, and enforced the generation of class-specific images. By generating images that are discriminable among classes, the discriminator learns consistent decision boundaries from both real and synthetic data, which in turn improves the classification performance. However, image data used in these studies are from general-purpose, high-quality and well-processed CV datasets, which may not reflect real conditions in practical scenarios.


In vision-based SHM, there exist a few early GAN studies \citep{gao2020peer, maeda2020generative}, and very little attention was directed towards the imbalanced-class issue. In addition, the proposed GAN-based pipeline in \citep{gao2019deep} for classification problems is shown to be computationally inefficient. According to the findings in relevant GAN-based classification studies \citep{salimans2016improved, madani2018semi}, the semi-supervised learning mechanism exploits the features of unlabeled data more thoroughly and simultaneously increases the model’s data generation and classification capabilities. Therefore, these findings have motivated us to convert the GAN into a semi-supervised variant.

Exploring GANs in engineering applications is still an open-ended discussion, and investigating GAN-based methods in vision-based SHM is the focus of this study. The main contributions of this work are:
\begin{itemize}

\item A simulative experimental configuration is designed. To simulate the restrictions of (1) low-data and (2) imbalanced-class, an extremely biased dataset containing limited images of undamaged state (UD), crack (CR) and spalling (SP) is built. To simulate the restriction of (3) limited computational resources, all pipelines are built on top of a shallow and general CNN classifier.

\item One of previously proposed GAN-based classification pipelines, namely synthetic data fine-tuning (SDF) \citep{gao2019deep}, is revisited with more heuristic reasoning.

\item A novel GAN-based classification pipeline with semi-supervised learning and balanced-batch sampling, namely the balanced-batch semi-supervised GAN (BSS-GAN), is proposed. 

\end{itemize}

Comparative computer experiments are conducted among BSS-GAN, baseline CNN (BSL), BSL using over-sampling, BSL with under-sampling, BSL with weighted loss, and other GAN-based methods. A key finding of this study is that BSS-GAN improves the damage detection performance in terms of recall and $F_\beta$ score and outperforms the above-mentioned methods under the low-data and imbalanced-class regime. The paper is organized as follows: Section 2 introduces the background of GAN and GAN-based augmentation methods; Section 3 describes the experimental details, dataset, evaluation metrics, and network configuration; Section 4 presents the experimental results and analysis, and finally, Section 5 contains the conclusions.

\section{GAN Background and GAN-based Augmentation Methods}
\subsection{Basics of GAN}
GAN is a generative model that generates new data from the learned distribution \citep{goodfellow2014generative}. Unlike conventional DL models, GAN consists of two networks, namely the \textit{generator} and the \textit{discriminator}, where the former creates synthetic data and the latter classifies an input sample as ``real'' or ``synthetic''. GAN uses adversarial training, where each network aims to minimize the gain of the opposite side while maximizing its own. Ideally, both the generator and the discriminator converge to the {\textit{Nash equilibrium}} \citep{goodfellow2016tutorial}, where the discriminator gives equal predictive probabilities to real and synthesized samples.

Suppose that $x \in \mathbb{R}^{d}$ is a sample in the $d$ dimensional space over the set of real numbers $\mathbb{R}$, $x_{r} \sim p_{\rm{data}}$ is a sample from the {\textit{real}} data distribution, and $x_g \sim p_g$ is a {\textit{generated}} sample from the GAN-learned synthetic data distribution. The generator $G$ with parameters $\theta_{G}$ is trained to synthesize samples that mimic $p_{\rm{data}}$ by mapping the noise vector (latent variable from distribution $p_z$), $z \sim p_{z}$, to a synthesized sample $x_g = G(z; \theta_{G})$. The discriminator $D$ with parameters $\theta_{D}$ takes in a sample $x$ (either $x_r$ or $x_g$) and outputs $D(x; \theta_D)$, which is the predictive probability that $x$ comes from $p_{\rm{data}}$ not $p_{g}$. For simplicity, $\theta_D$ and $\theta_G$ arguments are dropped in subsequent formulations. 

During training, $G$ competes with $D$ according to:
\begin{equation}
\label{eq_GAN_minimax}
\min_{G} \max_{D} \left\{E_{x \sim p_{\rm{data}}} \left[\ln D(x)\right] +
E_{z \sim p_{z}}\left[\ln(1 - D(G(z)))\right]\right\}
\end{equation}
where $E_{a\sim p_b}$ denotes the expectation of the data or the random variable $a$ sampled from the distribution $p_b$. The first term is the negated cross-entropy between $p_{\rm{data}}(x)$ and $D(x)$, whose value is positively associated with $D$'s ability of correctly predicting real samples from $p_{\rm{data}}(x)$, i.e., $x=x_{r}$; the second term is the negated cross-entropy between $p_{z}(z)$ and $1-D(G(z))$, which is $D$'s predictive probability that a synthesized sample $x_{g} = G(z)$ is indeed considered as ``synthetic'', i.e., $x=x_{g}$. $D$ aims to maximize its discriminative power characterized by both terms, while the generator $G$ tries to undermine $D$'s performance by synthesizing realistic samples to trick $D$ (minimizing the second term). Both $D$ and $G$ can be parametrized by deep neural networks or CNNs, and they are trained and optimized alternatively according to Equation (\ref{eq_GAN_minimax}) until reaching the optima or designated number of iterations.

\subsection{Synthetic data over-sampling}
\begin{figure}
\centering
\includegraphics[width=0.55\linewidth]{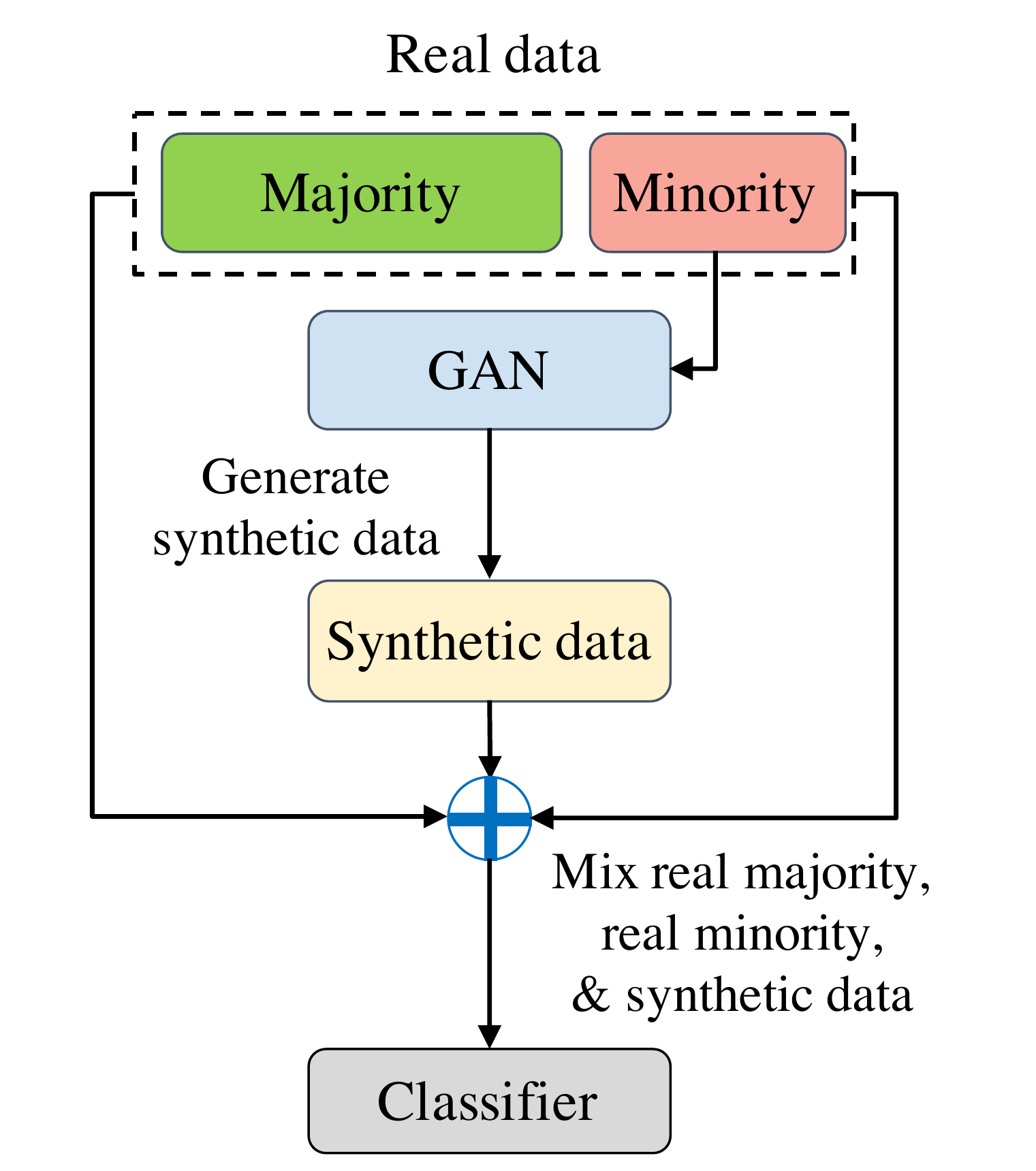}
\caption{\enspace Illustration of GAN-OS pipeline\label{fig_pipeline_GAN_OS}}
\end{figure}

GAN can be used to generate new data, which is thought to be useful in enriching the dataset. Therefore, one straightforward way is to over-sample the minority-class data by GAN-generated data to reduce the majority-to-minority ratio. This is named the GAN-based synthetic data over-sampling (GAN-OS), Figure \ref{fig_pipeline_GAN_OS}. However, preliminary investigations have demonstrated that such aggregation may sometimes render worse performance (e.g., more severe over-fitting) in the test set \citep{jain2018imagining, gao2019deep}. Possible reasons are: 1) relatively lower image quality compared to real ones, 2) inherent data biases, and 3) possible distribution differences between the synthetic and the real ones. In our opinion, the dominating failure factor is the false sense of building a series pipeline.

The generated data are thought of as fixed realizations of the generator network of the GAN model. In other words, the generated data are equivalent to a fixed-parameter generator. From Figure \ref{fig_pipeline_GAN_OS}, feeding a mixture of real and generated synthetic data into the classifier builds a series pipeline, i.e., a fixed-parameter generator followed by the original classifier along with the real data, where the number of parameters is the sum of those in the generator and the original classifier. Therefore, this series pipeline gains more parameters than the original classifier. Although the generated data increase the data size and possibly data variety, if the original classifier already suffers from over-fitting due to data deficiency, the additional parameters introduced to the series pipeline will increase the risk of further exacerbating over-fitting, such as the case in \citep{gao2019deep}.

\subsection{Synthetic data fine-tuning}
\begin{figure}
\centering
\includegraphics[width=0.6\linewidth]{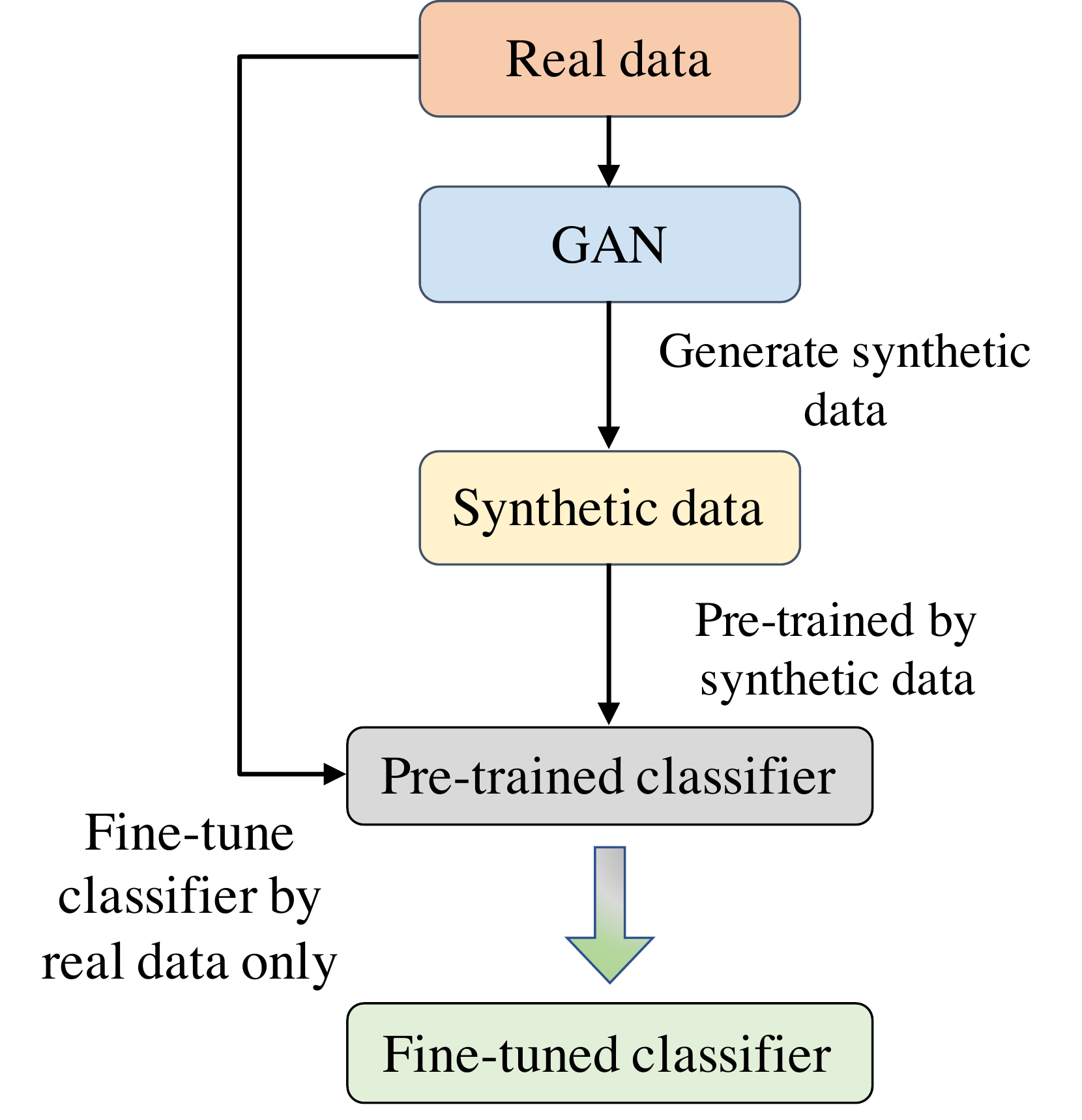}
\caption{\enspace Illustration of SDF pipeline\label{fig_pipeline_SDF}}
\end{figure}

\begin{figure*}[htbp]
\centerline{\includegraphics[width=0.7\linewidth]{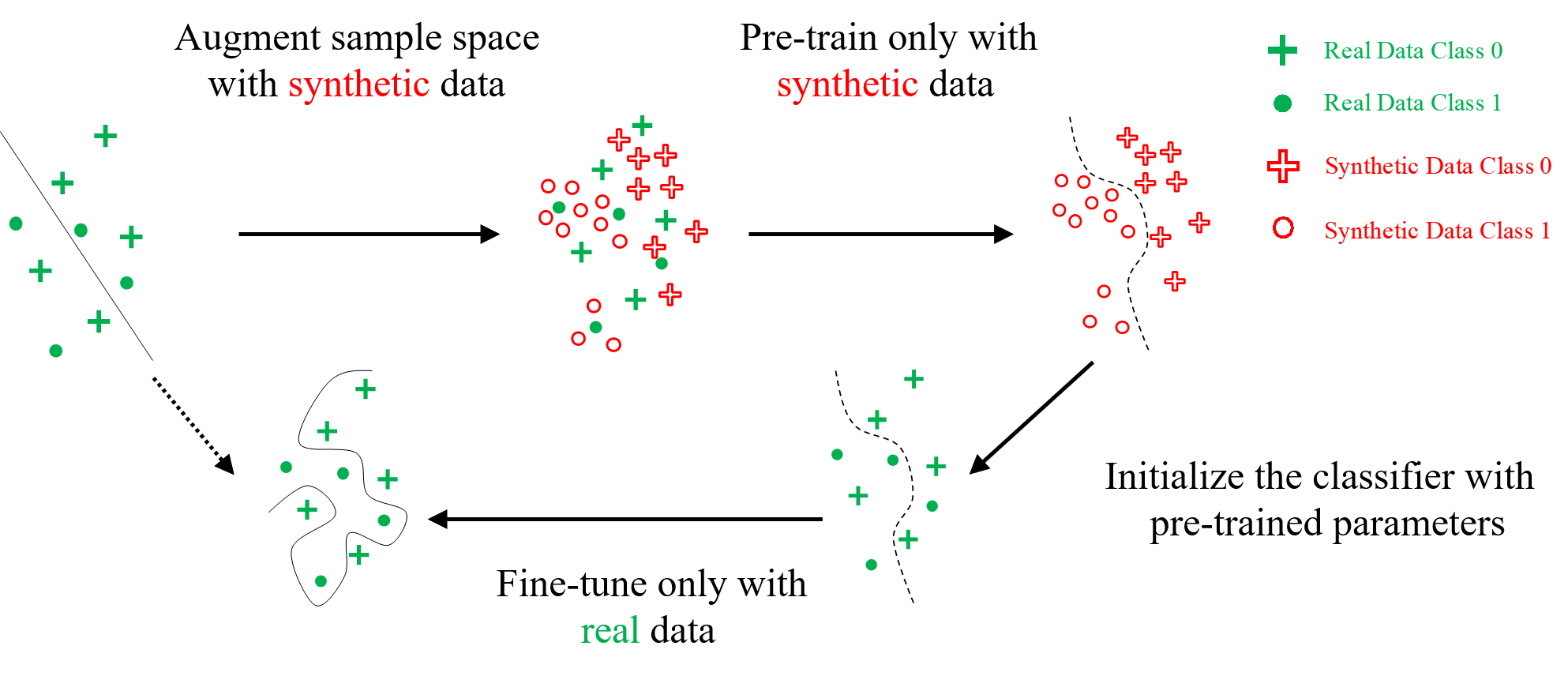}}
\caption{\enspace Two learning modes of conventional training and SDF\label{fig_intuition_SDF}}
\end{figure*}

To alleviate over-parametrization in GAN-OS, \cite{gao2019deep} proposed a pipeline based on TL, called synthetic data fine-tuning (SDF), where a weak classifier is firstly pre-trained on the generated synthetic images and then fine-tuned by the real ones, Figure \ref{fig_pipeline_SDF}. The intuition of this pipeline is explained in the following: using a weak or non-classic CNN classifier under computing power limitations usually implies that such a classifier does not have pre-trained parameters from ImageNet or other datasets in the source domain. Instead, its initialization fully depends on random, Gaussian, or other initialization approaches, e.g., Xavier initialization \citep{glorot2010understanding}, which may be poor due to the large random parameter space. Conceptually, it is difficult to learn the decision boundary well by directly training from such initialization as illustrated in Figure \ref{fig_intuition_SDF}. When considering the SDF pipeline, regardless of the correctness of the generated synthetic images, these images can be thought of as being generated from a similar or enlarged sample space, originating from the real raw data. Thus, it is believed that the classifier, when pre-trained from such enlarged space, provides a better initialization for the subsequent fine-tuning step with the real data.

\subsection{Balanced semi-supervised GAN}
The original GAN is trained in an unsupervised learning manner, and its discriminator only differentiates unlabeled real samples from those synthesized by the generator. Herein, a balanced semi-supervised GAN (BSS-GAN) is formulated in the manner of a semi-supervised learning \citep{salimans2016improved}. Unlike the ordinary GAN \citep{goodfellow2014generative}, the output dimension (size) of the discriminator increases from 2 (``real'' or ``synthetic'') to $K+1$ so that the discriminator can classify samples from $K$ real classes (samples follow $p_{\rm{data}}(x,y)$) and one ``synthetic'' class (generated samples follows $p_{g}$). Besides data $x$ and the corresponding label $y$ from $p_{\rm{data}}(x,y)$, the model also learns from unlabeled data distribution $p_{\rm{data}}(x)$, simultaneously. These characteristics illustrate the core concept of semi-supervised learning. We further introduce a balanced-batch sampling approach to pursue both class-balance and real-synthetic balance during training.

\subsubsection{Discriminator loss}
For each input sample $x$ taken from either $p_{\rm{data}}$ (from both $p_{\rm{data}}(x)$ and $p_{\rm{data}}(x,y)$) or $p_g$, the discriminator $D$ outputs a $(K+1)$-dimensional predictive probability vector $p_{\rm{model}}(y|x)$, where $p_{\rm{model}}(y=i|x)=\left.\exp D(x)_{i}\right/\sum_{j=1}^{K+1}\exp D(x)_{j}, \ i=1,...,K+1$, represents the predictive probability of the sample coming from the $i$-th class and $D(x)_{1}, \dots, D(x)_{K+1}$ are logits output by $D(x)$ corresponding to each class. Herein, $p_{\rm{model}}(y=K+1|x)$ represents the predicted probability that sample $x$ is “synthetic”, and thus $D(x)$ in Equation \ref{eq_GAN_minimax} can be substituted by $1-p_{\rm{model}}(y=K+1|x)$. Similarly, $1-D(G(z))$ in the second term of Equation \ref{eq_GAN_minimax} is equivalent to $p_{\rm{model}}(y=K+1|G(z))$. Both terms in Equation \ref{eq_GAN_minimax} are negated to form a minimization problem of $D$. The unsupervised (US) loss without using any label information of the $K$ real classes is derived as:
\begin{eqnarray}
\label{eq_unsupervised_loss}
L^{(D)}_{\rm{US}} & = & - E_{x \sim p_{\rm{data}}} \ln(1 - p_{\rm{model}}(y=K+1|x)) \nonumber \\
                  &   & - E_{x \sim p_{g}} \ln(p_{\rm{model}}(y=K+1|x))
\end{eqnarray}
Therefore, both labeled and unlabeled data can be used for unsupervised feature learning in Equation \ref{eq_unsupervised_loss}.

For real labeled data, the supervised (S) discriminator loss is the cross-entropy between the real labeled data distribution $p_{\rm{data}}(x,y)$ and the model's predicted label distribution for $K$ real classes (given real input sample $x$), $p_{\rm{model}}(y|x, y < K+1)$:
\begin{equation}
\label{eq_supervised_loss}
L^{(D)}_{\rm{S}} = - E_{x,y \sim p_{\rm{data}}(x,y)} \ln(p_{\rm{model}}(y|x, y<K+1))
\end{equation}
Finally, the total discriminator loss is expressed as follows:
\begin{equation}
\label{eq_D_loss}
L^{(D)} = L^{(D)}_{\rm{US}} + L^{(D)}_{\rm{S}}
\end{equation}

\subsubsection{Generator loss}
The generator's objective is to ``weaken'' the discriminator's performance. However, the original formulation of the generator loss \citep{goodfellow2014generative} usually does not perform well. This is because the generator's gradient vanishes when the discriminator has high confidence of distinguishing the generated samples from the real ones, i.e., when $D(G(z)) \rightarrow 0$. Therefore, the generator loss used in our formulation refers to the heuristic (H) loss \citep{goodfellow2016tutorial} as follows,
\begin{equation}
\label{eq_heuristic_G_loss}
L^{(G)}_{\rm{H}} = -\frac{1}{2}E_{z\sim p_{z}} \ln{D(G(z))}
\end{equation}

Instead of minimizing the expected ln-probability of the discriminator being correct, the generator now maximizes the expected ln-probability of the discriminator making a mistake, i.e., assigning a  ``real'' label to a generated sample $G(z)$. In this study, the constant multiplier in Equation (\ref{eq_heuristic_G_loss}), i.e., $1/2$, is dropped and $E_{z \sim p_z} \ln{D(G(z))}$ is substituted by $E_{z \sim p_z} \ln{[1-p_{\rm{model}}(y=K+1|G(z))]}$ to accommodate the $(K+1)$-dimensional discriminator output, i.e.,
\begin{equation}
\label{eq_SSGAN_heristic_loss}
L^{(G)}_{\rm{H}} = -E_{z \sim p_z} \ln[1-p_{\rm{model}}(y=K+1|G(z))]
\end{equation}

Feature matching (FM) is a technique that prevents over-training the generator and increases the stability of the GAN \citep{salimans2016improved}. It requires the generator to produce samples which result in similar features on an intermediate layer of the discriminator network as do the real samples. Thus, the generator loss considering FM is formulated as follows,
\begin{equation}
\label{eq_G_loss_fm}
L^{(G)}_{\rm{FM}} = \Big\Vert E_{x \sim p_{\rm{data}}} f(x)-E_{z\sim p_{z}} f(G(z)) \Big\Vert_{2}^{2}    
\end{equation}
where $||\bullet||_2$ is the ${\cal{L}}_2$ norm of $\bullet$, $f(x)$ is the activation of an intermediate layer of the discriminator for sample $x$. In this study, $f(x)$ is defined by the ReLU \citep{relu2010} activation on the flattened output of the last convolutional (Conv) layer of the discriminator network. Finally, combining $L^{(G)}_{\rm{H}}$ and $L^{(G)}_{\rm{FM}}$, the total generator loss is expressed as follows,
\begin{equation}
\label{eq_G_loss}
L^{(G)} = L^{(G)}_{\rm{H}}+L^{(G)}_{\rm{FM}}
\end{equation}

Besides cross-entropy based loss formulations (Equations \ref{eq_D_loss} \& \ref{eq_heuristic_G_loss}) used in this study, other loss functions, e.g., Wasserstein GAN loss (\cite{arjovsky2017wasserstein}), Least Squares GAN loss (\cite{mao2017squares}) can also be modified to fit the purpose of semi-supervised GAN training, which is worth future investigations.  

\subsubsection{Balanced-batch sampling}
The class-imbalance issue not only affects the classification performance, but also deteriorates the perceptual quality and diversity of the generated samples \citep{mariani2018bagan}. During the conventional training (updating) procedure of a DL classifier or GAN, \textit{mini-batch gradient descent} is commonly adopted \citep{Goodfellow2016DL}. Due to computational limitations, instead of using {\textit{all}} the data at once, the DL network is only fed with {\textit{one small batch}} containing $m$ data samples randomly selected from the full dataset of size $N$, where $m$ is called the batch size and $m < N$. Statistically, if the dataset is imbalanced, the batch is also imbalanced, which eliminates neither the performance bias of the classifier nor the training instability of GAN.


In this paper, we introduce the \textit{balanced-batch sampling} in training. As its name suggests, while forming the batch, the same amount of data is randomly sampled from each class. For balanced-batch sampling in GAN training, two types of balances are maintained in a given batch: (1) balance among real classes in the labeled data, and (2) balance between any particular real class and the ``synthetic'' class. For (1), $n_{l}$ real labeled samples are randomly selected from $K$ real classes, where $n_{1}^{l},n_{2}^{l},\dots,n_{K}^{l}$ are the numbers of data from each class:

\begin{equation}
\label{eq_equal_class}
n_{1}^{l} = n_{2}^{l} = \dots = n_{K}^{l} = n_{l}/K
\end{equation}

For (2), the total amount of generated samples $n_{g}$ matches any of the sub-batches from a certain real class, i.e., $n_{g} = n_{l}/K =n_{k}^{l}, \forall k \in \{1,2,\dots,K\}$. In other words, each of the $K+1$ classes contributes a sub-batch of the same size to the whole batch:

\begin{equation}
\label{eq_batch_formation}
m = n_{l} + n_{g} = (K+1) \ n_{l}/K
\end{equation}

The BSS-GAN can utilize the unlabeled data in feature learning, Equations \ref{eq_unsupervised_loss} \& \ref{eq_G_loss_fm}. If additional unlabeled data are available, in any given batch, the ratio of unlabeled samples to any single-class sub-batch is controlled by a hyper-parameter $c$, i.e., $n_{ul}/n_{l} = c/K$. In this study, to keep a smaller batch size to simulate computational limitations, $c=1$ is used. Therefore, the general formation for each batch is expressed in Equation \ref{eq_batch_formation_general}. 

\begin{equation}
\label{eq_batch_formation_general}
m = n_{l} + n_{g} + n_{ul} = (K+2) \ n_{l}/K
\end{equation}

\subsubsection{BSS-GAN algorithm}
\begin{figure*}[htbp]
\centerline{\includegraphics[width=0.8\linewidth]{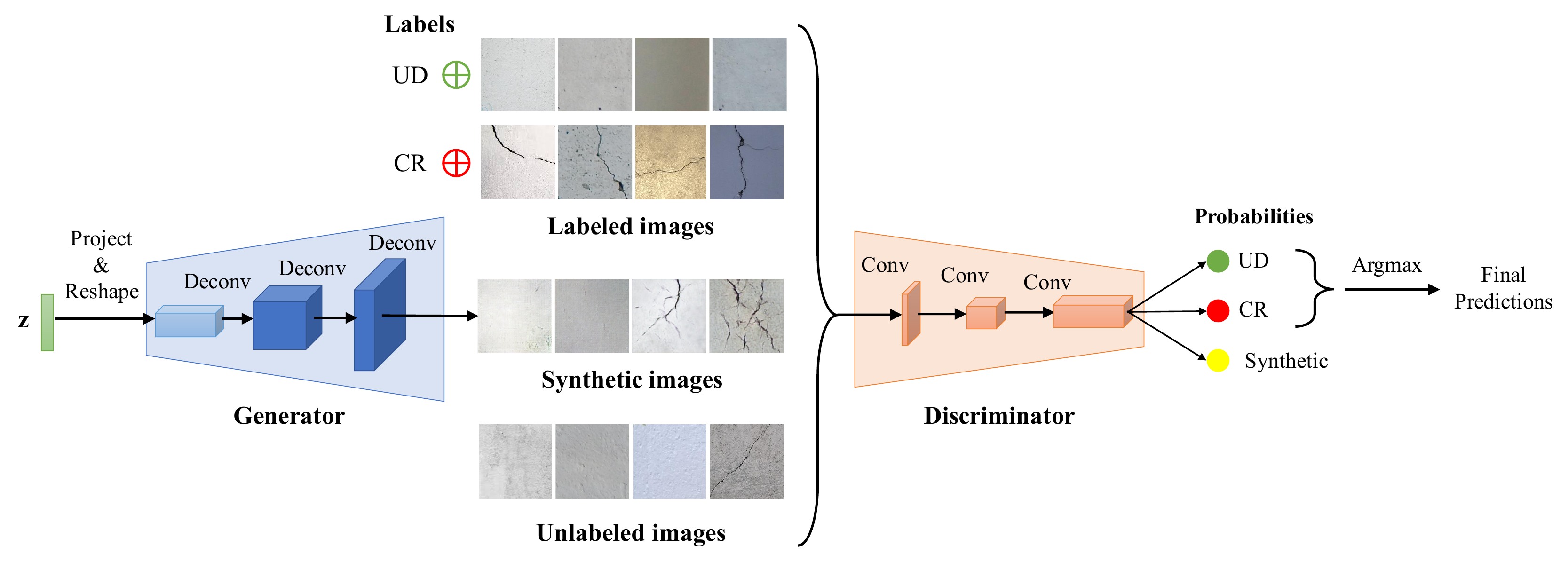}}
\caption{\enspace BSS-GAN pipeline in concrete crack detection (UD: undamaged; CR: cracked)\label{fig_BSSGAN_Framework}}
\end{figure*}

By formulating the GAN in a semi-supervised learning setting and using the balanced-batch sampling technique, we obtain the proposed integrated model BSS-GAN. This model builds an end-to-end pipeline for both synthetic image generation and classifier training, and it is expected to have a stable and less biased performance under highly imbalanced datasets. Moreover, unlike training a supervised learning-based DL classifier, unlabeled data are put into use. One example of using BSS-GAN in concrete crack detection is illustrated in Figure \ref{fig_BSSGAN_Framework}.

For a training batch size $m$, the detailed training procedure of the BSS-GAN is as follows:
\begin{enumerate}[align=left, start=0, label={Step \arabic*:}]
\item Initialize the discriminator $D$ and the generator $G$ with $\theta_{D}$ and $\theta_{G}$, respectively.
\item A subset of the batch represented by real data (both labeled and unlabeled) is formed, $B_{r} = B_r^{l} \bigcup B_r^{ul}= \{(x_{1}^{l},y_{1}^{l}),\dots,(x_{n_{l}}^{l},y_{n_{l}}^{l})\} \bigcup  \{x_{1}^{ul}, \dots, x_{n_{ul}}^{ul}\}$, where $n_{l}$ data-label pairs are equally sampled from the $K$ real classes, and class label $y_{i}^{l} \in \{1,2,\dots,K\}$.
\item Random noise vectors, $z = \{ z_{1}, \dots, z_{n_{g}}\}$, are sampled from the noise prior distribution $p_{g}(z)$ where $z$ is fed to $G$ to generate $n_{g}$ synthetic samples, $B_{g} = \{ G(z_{1}), \dots, G(z_{n_{g}})\}$, the remaining subset of the batch.
\item Feed $B_{r}$ to $D$ where for $x_{i} \in B_{r}$, $i \in \{1, \dots, n_{l}+n_{ul}\}$, $D$ outputs a $(K+1)$-dimensional probability vector, $u_{i} = [p_{\rm{model}}(y=1|x_{i}), \dots,  p_{\rm{model}}(y=K+1|x_{i})]^{T}$, where $u^{K+1}_i = p_{\rm{model}}(y=K+1|x_{i})$.
\item Feed $B_r^l$ to $D$ where for $(x_{i}^{l}, y_{i}^{l}) \in B_r^l, i \in \{1, \dots, n_{l}\}$, $D$ outputs a $(K+1)$-dimensional probability vector and only the first $K$ dimensions are considered,  $v_{i} = [p_{\rm{model}}(y=1|x_{i}^{l}), \dots, p_{\rm{model}}(y=K|x_{i}^{l})]^{T}$, where $v_{i}({y_{i}^{l}})=p_{\rm{model}}(y=y_{i}^{l}|x_{i}^{l})$ represents the probability of the model for predicting class $y_{i}^{l}$.
\item Feed $B_{g}$ to $D$ where for $G(z_{i}) \in B_{g}, i \in \{ 1, \dots, n_{g}\}$, $D$ outputs a $(K+1)$-dimensional probability vector $w_{i}=[p_{\rm{model}}(y=1|G(z_{i})), \dots, p_{\rm{model}}(y=K+1|G(z_{i}))]^{T}$, where $w^{K+1}_i = p_{\rm{model}}(y=K+1|G(z_{i}))$.
\item Compute the discriminator loss, $L^{(D)}$:

\begin{eqnarray}
\label{eq_D_loss_batch_sample}
L^{(D)} & = & -\frac{1}{n_{l}+n_{ul}}\sum_{i=1}^{n_{l}+n_{ul}} \ln\left(1- u^{K+1}_i\right) \nonumber \\
        &   & - \frac{1}{n_{l}}\sum_{i=1}^{n_{l}} \ln\left(v_i(y_{i}^{l})\right) -\frac{1}{n_{g}}\sum_{i=1}^{n_{g}} \ln\left(w^{K+1}_i\right) \hspace{5 mm}
\end{eqnarray}

\item Compute the generator loss, $L^{(G)}$:

\begin{eqnarray}
\label{eq_G_loss_batch_sample}
L^{(G)} & = & \Big\Vert \frac{1}{n_{l}+n_{ul}}\sum_{i=1}^{n_{l}+n_{ul}}f(x_{i}) - \frac{1}{n_{g}} \sum_{i=1}^{n_{g}} f\left(G(z_{i})\right) \Big\Vert_{2}^{2} \nonumber \\
        &   & -\frac{1}{n_{g}}\sum_{i=1}^{n_{g}} \ln\left(1-w^{K+1}_i\right) \hspace{3 cm} 
\end{eqnarray}

\item Optimize and update the network parameters $\theta_{D}$ and $\theta_{G}$, where $\eta$ is the learning rate.
\begin{equation}
\theta_{D} \leftarrow \theta_{D} - \eta \nabla_{\theta_{D}} L^{(D)} \hspace{.2 cm} \& \hspace{.2 cm} \theta_{G} \leftarrow \theta_{G} - \eta \nabla_{\theta_{G}} L^{(G)}
\end{equation}
\end{enumerate}

\noindent Repeat steps (1) to (8) until convergence is achieved or the designated number of iterations is reached. 


Balanced-batch sampling (steps 1 to 5) is only adopted in training. Once the BSS-GAN is trained, when referencing or predicting new data for classification purposes, all new data are fed into $D$ only, and the predicted class is the one with the highest predictive probability among the first $K$ entries of the $D$'s output. Similarly, for synthetic data generation, the noise vector $z$ is sampled and fed into $G$, which then outputs the synthetic data.

The traditional GAN-based over-sampling approach, e.g., GAN-OS, is offline, where synthetic data are generated and stored on the hard-disk before the training starts. Moreover, the number of generated data is pre-determined and fixed. On the contrary, BSS-GAN is an online DA process, where only $n_g$ synthetic data are generated and utilized during each training batch, and can be discarded when training moves to the next batch. The total amount of the generated data depends on the number of training batches. For example, when training moves to the $E$-th epoch, the total amount of the generated data is $n_g \cdot E \cdot \lceil N_l/n_l \rceil$, where $N_l$ is the total amount of the labeled data and $\lceil \cdot \rceil$ is the ceiling operation.

\section{Experimental preparation}
\subsection{Experimental objectives}
\begin{figure}
\centering
\begin{subfigure}{.31\linewidth}
  \centering
  \includegraphics[width=\linewidth]{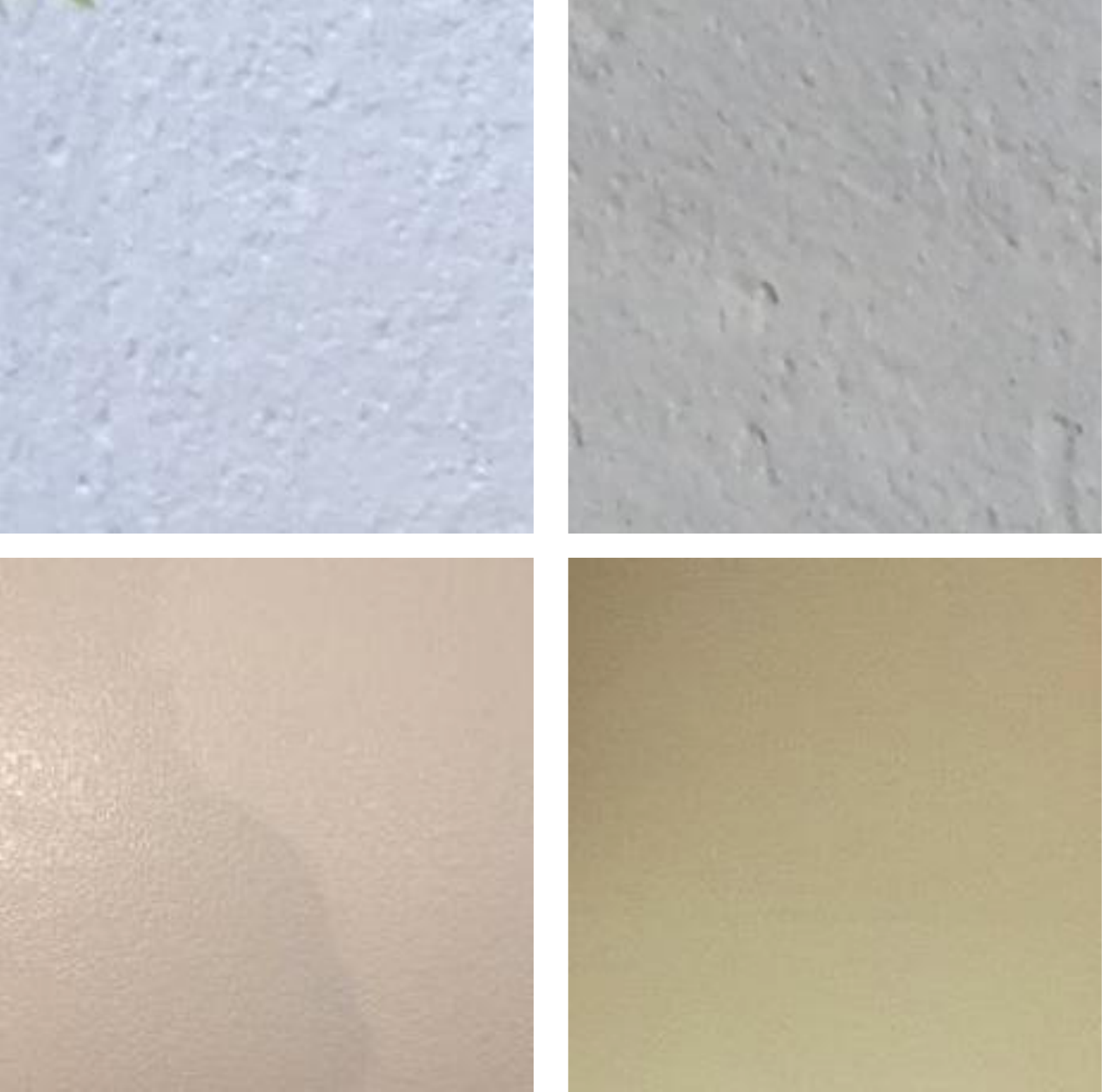}
  \caption{\label{fig_samples_a}Undamaged}
\end{subfigure} \hspace{1pt}
\begin{subfigure}{.31\linewidth}
  \centering
  \includegraphics[width=\linewidth]{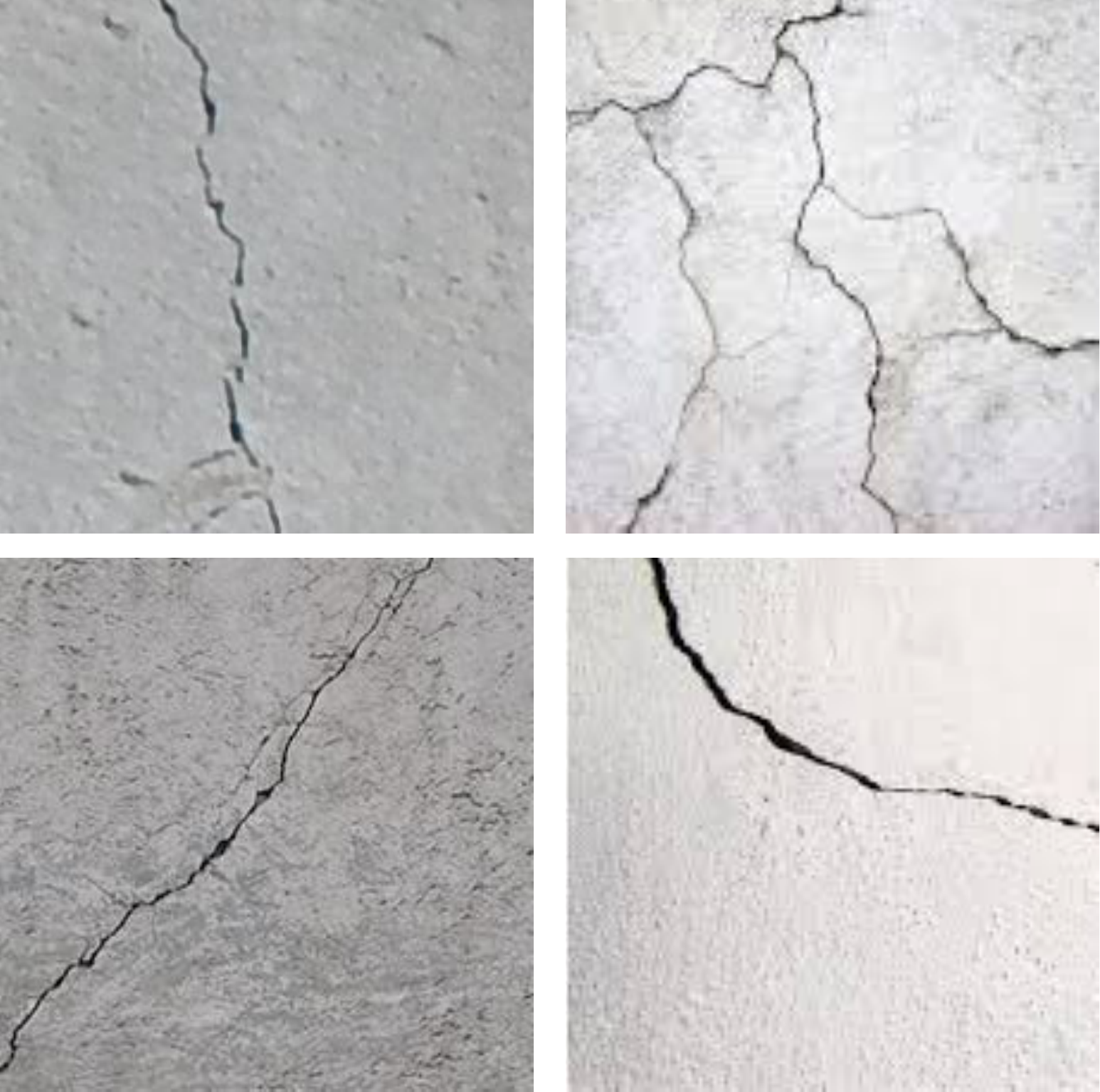}
  \caption{\label{fig_samples_b}Cracked}
\end{subfigure} \hspace{1pt}
\begin{subfigure}{.31\linewidth}
  \centering
  \includegraphics[width=\linewidth]{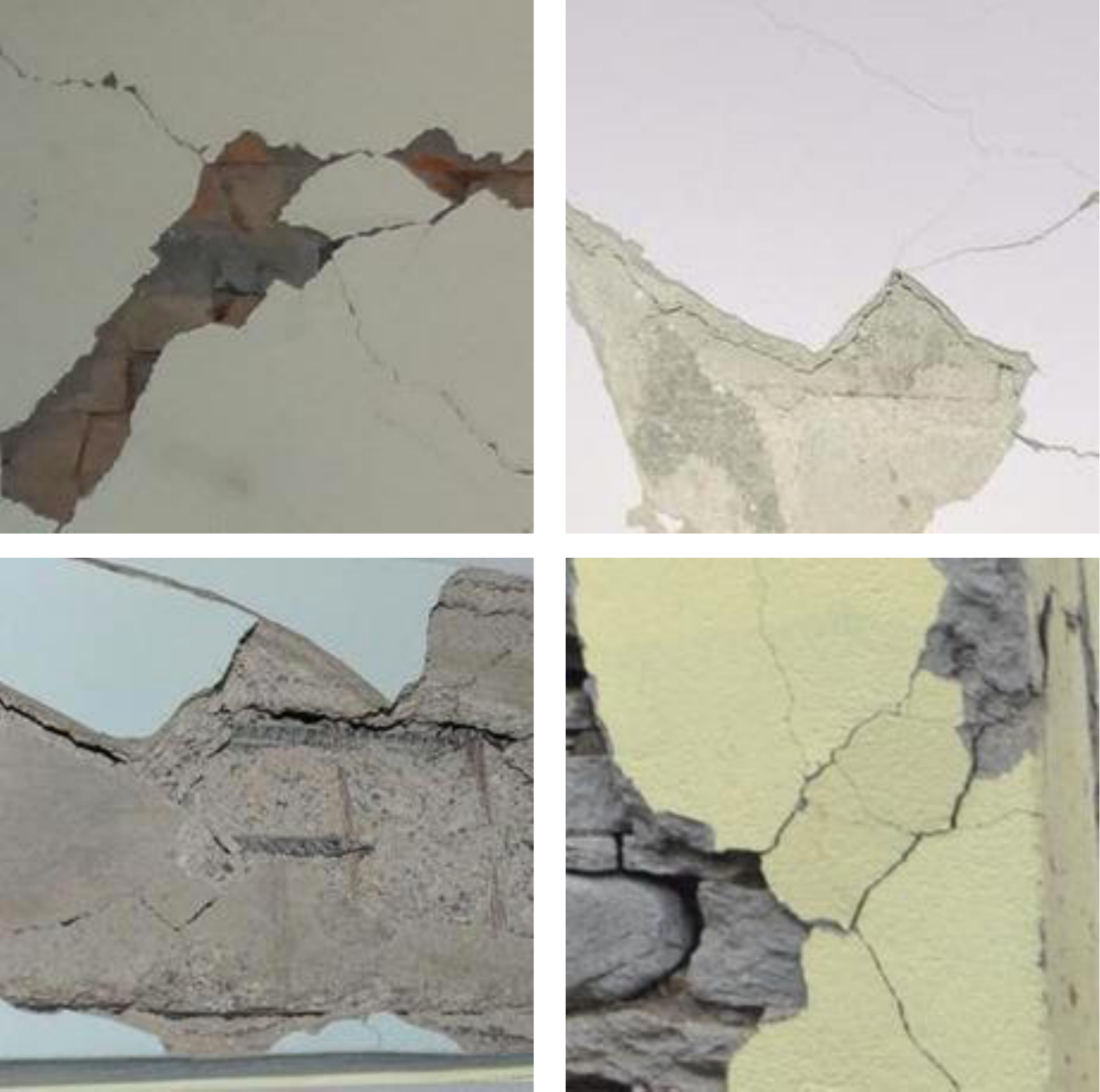}
  \caption{\label{fig_samples_c}Spalling}
\end{subfigure}
\caption{\label{fig_sample_images}\enspace Sample images of three classes}
\end{figure}

In the subsequent computer experiments, we aim to examine BSS-GAN's performance under low-data and imbalanced-class regimes on one common task in vision-based SHM, namely reinforced concrete damage detection. Three key statuses: (1) undamaged state (UD), Figure \ref{fig_samples_a}, (2) cracked (CR), Figure \ref{fig_samples_b}, and (3) spalling (SP), Figure \ref{fig_samples_c}, are considered, describing three damage levels in the order of increasing corrosion risk to embedded reinforcing bars. In real-world applications, the class ratios of UD:CR and UD:SP are usually high. To simulate such imbalances in a realistic SHM data collection, an empirical ratio of 32:2:1 (UD:CR:SP) is selected for experimental purposes, where we also treat SP as less frequent than CR. It is noted that the main purpose of this study is to circumvent the high imbalance issue in practice, thus, slightly imbalanced or even balanced class ratios are not discussed herein. Three major validation experiments are designed:
\begin{enumerate}
\item Binary crack detection with UD:CR class ratio of 16:1.
\item Binary spalling detection with UD:SP class ratio of 32:1.
\item Ternary damage pattern classification with UD:CR:SP class ratio of 32:2:1.
\end{enumerate}
Experiments (1) \& (2) simulate the real-world binary damage detection in vision-based SHM, where the number of ``undamaged'' cases (UD) far exceeds that of ``damaged'' cases (CR or SP). On the other hand, experiment (3) integrates the two damage cases CR \& SP into a comprehensive but more complex three-class classification, which aims to evaluate the DL models in an imbalanced multi-class problem.


For a comparative study, in each experiment, six pipelines are configured and compared as follows:

\begin{enumerate}
\item A baseline shallow CNN classifier (BSL).
\item BSL with under-sampling the majority-class data to restore the class balance (BUS).
\item BSL with over-sampling minority-class data by conventional DA, e.g., flip, translation, and rotation (BOS-DA).
\item BSL with over-sampling the minority-class data by ordinary GAN-generated data (BOS-GAN).
\item BSL adopting the SDF training pipeline (BSL-SDF).
\item BSS-GAN.
\end{enumerate}

In BUS, BOS-DA \& BOS-GAN pipelines, an equal class ratio is maintained after under-/over-sampling. The performance of each case is evaluated by appropriate metrics, e.g., recall, confusion matrix, and $F_\beta$ score, which are covered in more details in the following subsection. Beyond these, more intuitions are discussed in terms of (i) synthetic image quality of ordinary GAN and BSS-GAN, (ii) effectiveness of unsupervised feature learning using different amounts of unlabeled data in BSS-GAN, and (iii) comparison between BSS-GAN and BSL using weighted loss function methods. Points (ii) \& (iii) are explored in the extended experiments (4) \& (5). 

\subsection{Dataset}
For generality, the images in this study were obtained from two open-source structural image datasets: PEER Hub ImageNet ($\phi$-Net) \citep{gao2020peer} and SDNET2018 \citep{dorafshan2018comparison}. The structural images cover scenarios ranging from undamaged to extreme cracking or spalling. The images were further processed for the experiments as follows:
\begin{enumerate}
\item Clean the dataset and select pixel-level (close up) images with high visual quality.
\item Select and store the images with labels UD, CR, and SP to build the full dataset.
\item Rescale the images to the uniform size of 128$\times$128 pixels using bicubic resampling.
\end{enumerate}

Finally, a dataset with a total of 15,750 images was constructed including 14,400 UD, 900 CR, and 450 SP images, Table \ref{tab_dataset_statistics}. In addition, a 2:1 training-test split ratio was applied, compared with common 3:1 or 4:1 ratios. This led to fewer training data, which also helped simulate the data deficiency. On the other hand, in this manner, there were sufficient test data (especially minority-class data) to be appropriately evaluated using the proposed BSS-GAN.
Compared with the $\phi$-Net benchmark experiments \citep{gao2020peer} and general CV applications, this dataset is imbalanced and low-data. In addition, the dataset is publicly available upon request from \url{http://stairlab.berkeley.edu/data/}.

\begin{table}[ht]
\centering
\caption{\enspace Label statistics of the experimental dataset}
\label{tab_dataset_statistics}
\vspace*{-1mm}
\begin{tabular}{c|c|c|c|c}
\hline
\textbf{Label} & \textbf{UD}     & \textbf{CR}    & \textbf{SP}    & \textbf{Class ratio} \\ \hline
Training       & 9,600   & 600   & 300   & 32:2:1      \\ \hline
Test           & 4,800   & 300   & 150   & 32:2:1      \\ \hline
Total          & 14,400  & 900   & 450   & 32:2:1      \\
\hline
\end{tabular}
\end{table}

To investigate the contribution of unlabeled data in BSS-GAN, an imbalanced ``hybrid'' dataset for concrete crack detection (UD vs. CR) was formed with 20\% of the labeled training data ($0.2\times(9,600+600)=2,040$) in the original crack detection, which simulated a more severe low-data scenario. The remaining training data ($0.8\times(9,600+600)=8,160$) were treated as unlabeled. Both labeled and unlabeled data remained at UD:CR=16:1 class ratio.

  
  
  

\subsection{Evaluation metrics}
Classification (overall) accuracy is defined by the number of correct predictions divided by the total number of predictions made for a dataset. This is not informative for imbalanced classification problems where merely guessing all samples from the majority class yields a misleadingly high accuracy. Thus, in this study, appropriate metrics are introduced and used. 

\subsubsection{Confusion matrix}
For classification, confusion matrix (CM) is a useful tool to summarize the model performance. For a binary classification problem, the CM has four possible outcomes: true positive (TP), true negative (TN), false positive (FP), and false negative (FN). By normalizing the CM entries with the number of predictions of each class, the diagonal entries become the true positive rate (TPR) and the true negative rate (TNR), Equations \ref{eq_TPR_TNR}, which are used for evaluating the accuracy of detecting TP and TN outcomes, respectively. In addition, the normalized CM can be applied to multi-class problems, i.e., $K > 2$, for which the recall (defined in the next paragraph) for each class is evaluated and placed on the diagonal cells.

Recall\footnote{In the binary case, recall is usually defined by the accuracy of correctly predicting the TP outcome, Equation \ref{eq_precision}, which is equivalent to TPR Equations \ref{eq_TPR_TNR}.} and precision, Equation \ref{eq_precision}, are other commonly used metrics. Recall/TPR and precision measure different aspects of a model's performance where the former and latter are more appropriate for minimizing FN and FP, respectively \citep{chawla2004special}. In damage detection, damaged and undamaged states are usually defined as positive and negative, respectively. Thus in this study, the number of negative data far exceeds the number of positive ones. Failing to detect damages (i.e., more FN) bears severer consequences than wrongly recognizing undamaged samples as damaged (i.e., more FP). Accordingly, the first focus of the trained model should be to minimize the FN, measured by recall/TPR. On the other hand, precision is not an appropriate metric because due to the large number of negative (undamaged) data, a small drop in TNR will cause a large increase in FP, which overwhelms the TP and leads to a misleadingly low precision value. Furthermore, TNR can be an alternative to recall/TPR, which takes FP into account and measures the accuracy of the TN detection.

\begin{equation}
\mathrm{TPR} 
=  \dfrac{\mathrm{TP}}{\mathrm{TP} + \mathrm{FN}} \hspace{5 mm} \& \hspace{5 mm} \mathrm{TNR} =  \dfrac{\mathrm{TN}}{\mathrm{TN} + \mathrm{FP}}
\label{eq_TPR_TNR}
\end{equation}

\begin{equation}
\mathrm{Recall} =  \dfrac{\mathrm{TP}}{\mathrm{TP} + \mathrm{FN}} \hspace{5 mm} \& \hspace{5 mm} 
\mathrm{Precision} =  \dfrac{\mathrm{TP}}{\mathrm{TP} + \mathrm{FP}} \label{eq_precision}
\end{equation}

\subsubsection{$F_\beta$ score}
Besides CM, the $F_\beta$ score is another suitable metric for imbalanced binary classification problems \citep{chawla2004special}. Instead of completely ignoring the precision, $F_\beta$ weights and combines both precision and recall scores into a single measurement, Equation \ref{eq_fbeta}. Based on different $\beta$ values, $F_\beta$ measures varying importance of recall over precision. When $\beta=1$, both recall and precision are weighted equally ($F_1$ score).

The $\beta$ factor has real-world interpretations, and using F$_\beta$-$\beta$ graph is easy to interpret and communicate to engineers or stakeholders \citep{chawla2004special}, e.g., a larger $\beta$ can be treated as a measure that emphasizes the cost of mislabeling a damaged sample compared to producing a false alarm. According to \citep{chawla2004special}, $\beta=2$ is a common value ($F_2$ score). However, in SHM applications, the low tolerance of missing a damage calls for a higher $\beta$. In this study, we consider $\beta=2$ ($F_2$), $\beta=5$ ($F_5$), and other $\beta$ values from 1 to 10.

\begin{equation}
F_{\beta} = (1 + \beta^2) \frac{\mathrm{Precision} \times \mathrm{Recall}}{(\beta^2 \times \mathrm{Precision}) + \mathrm{Recall}} \label{eq_fbeta}
\end{equation}


\subsection{Network configurations}
In many previous studies, the network design is sophisticated and task/dataset dependent. To avoid loss of generality while considering the possible computing power limitations in field applications, the BSL is designed as a general multi-layer CNN discriminative classifier with a simple network configuration and hyper-parameter tuning, Table \ref{tab_config_discriminator}. For a fair comparison, the BSS-GAN's discriminator uses the same architecture as the BSL, e.g., same numbers of layers and filters, except for the output dimension ($C=K+1$ in BSS-GAN as opposed to $C=K$ in the BSL). According to \citep{radford2015unsupervised}, Leaky ReLU \citep{maas2013rectifier} is used as the activation function with a negative slope coefficient $\alpha=0.2$, and batch normalization (BatchNorm) \citep{ioffe2015batch} layers are inserted after the intermediate Conv layers with momentum 0.8. To avoid over-fitting, a 0.25 dropout rate is also applied.

The generator of the BSS-GAN is configured based on previous studies \citep{ radford2015unsupervised, gao2019deep}, Table \ref{tab_config_generator}. Due to the small image size (128$\times$128), a conventional 100-dimensional noise vector is randomly generated from the Gaussian distribution as the input to the generator without dropout \citep{srivastava2014dropout}. BatchNorm layers with a 0.8 momentum are added after the deconvolutional (Deconv) layers except for the last layer.

For the other two GAN-based pipelines (BOS-GAN \& BSL-SDF), the GAN portions are consistent with that of the BSS-GAN. The exception is the loss function, which is taken as the one used in the original GAN \citep{goodfellow2014generative}.

\begin{table*}[htbp]
\centering
\caption{\enspace Configurations of the BSL or the discriminator of the BSS-GAN}
\label{tab_config_discriminator}
\vspace*{-1mm}
\begin{tabular}{c|c|c|c|c}
\hline
\textbf{Layer} & \textbf{Filter size (\#)} & \textbf{Activation} & \textbf{Shape} & \textbf{Notes ($\alpha$: $-$ive slope coef. in Leaky ReLU)}\\
\hline
Input     &         -         &        -       &   ($N$, 128, 128, 3)    &   Input RGB images of size 128$\times$128    \\\hline
Conv      &   3$\times$3 (32)  &    Leaky ReLU  &   ($N$, 64, 64, 32)     &    Stride = 2, $\alpha=0.2$          \\\hline
Dropout   &         -         &        -       &   ($N$, 64, 64, 32)     &    Dropout rate = 0.25               \\\hline
Conv      &   3$\times$3 (64)  &    Leaky ReLU  &   ($N$, 32, 32, 64)     &    Stride = 2, $\alpha=0.2$          \\\hline
BatchNorm &         -         &        -       &   ($N$, 32, 32, 64)     &    Momentum = 0.8                    \\\hline
Dropout   &         -         &        -       &   ($N$, 32, 32, 64)     &    Dropout rate = 0.25               \\\hline
Conv      &   3$\times$3 (64)  &    Leaky ReLU  &   ($N$, 32, 32, 64)     &    Stride = 1, $\alpha=0.2$          \\\hline
Flatten   &         -         &        -       &   ($N$, 65,536)         &    65,536 = 32$\times$32$\times$64     \\\hline
Fc-layer  &         -         &     Softmax    &   ($N$, $C$)            &    BSS-GAN: $C=K+1$; BSL: $C=K$    \\
\hline
\end{tabular}
\end{table*}

\begin{table*}[htbp]
\centering
\caption{\enspace Configuration of the generator of the BSS-GAN}
\label{tab_config_generator}
\vspace*{-1mm}
\begin{tabular}{c|c|c|c|c}
\hline
\textbf{Layer} & \textbf{Filter size (\#)} & \textbf{Activation} & \textbf{Shape} & \textbf{Notes}\\
\hline
Input       &         -          &        -       &   ($N$, 100)              &    Noise generated from Gaussian distribution   \\\hline
Fc-layer    &         -          &       ReLU     &   ($N$, 131,072)          &    131,072 = 32$\times$32$\times$128    \\\hline
Reshape     &         -          &        -       &   ($N$, 32, 32, 128)      &     - \\\hline
Deconv      &   3$\times$3 (128)  &       ReLU     &   ($N$, 64, 64, 64)       &    Stride = 2          \\\hline
BatchNorm   &         -          &        -       &   ($N$, 64, 64, 64)       &    Momentum = 0.8      \\\hline
Deconv      &   3$\times$3 (64)   &       ReLU     &   ($N$, 128, 128, 3)      &    Stride = 2          \\\hline
BatchNorm   &         -          &        -       &   ($N$, 128, 128, 3)      &    Momentum = 0.8      \\\hline
Deconv      &   3$\times$3 (3)    &       Tanh     &   ($N$, 128, 128, 3)      &    Stride = 1          \\
\hline
\end{tabular}
\end{table*}

\subsection{Experimental setups}
In the three experiments discussed in Section 4.1, all data were labeled. For the first five models, also listed in Section 4.1, a batch size of $m=60$ was used. For the the sixth model of BSS-GAN, the number of labeled real data was maintained at 60, for the remaining batch data, based on balanced-batch sampling, the numbers varied for different cases. For the binary cases and the three-class case, the total batch size was $m = 60 + 60/2 = 90$ and $m = 60 + 60/3 = 80$, respectively. The fourth experiment with unlabeled data for binary crack detection had $m = 60 + 60/2 \times 2 = 120$.

To compare with the weighted loss function methods in the fifth experiment, two common weighted loss functions, namely balanced cross entropy ($L_{BCE}$), Equation \ref{eq_loss_bce}, and focal loss ($L_{F}$), Equation \ref{eq_loss_focal}, are individually used in the BSL.

\begin{equation}
L_{BCE} =  \sum_{i=1}^{K} -\alpha_i \ t_i \ln(p_i)
\label{eq_loss_bce}
\end{equation}

\begin{equation}
L_{F} = \sum_{i=1}^{K} -\alpha_i \ (1-p_i)^{\gamma} \ t_i \ln(p_i)
\label{eq_loss_focal}
\end{equation}

\noindent where $\alpha_i>0, i\in\{1,\dots,K\}$, is the class-weighting factor, $p_i, i\in\{1,\dots,K\}$ is the predicted probability for the $i$-th class, $t_i$ represents the label information, i.e., $t_i = 1$ when $i$ equals to the class label $y$, otherwise $t_i = 0$, and $\gamma \geq 0$ is the ``focusing parameter'', which down-weights the loss contribution from easy examples with high $p_i$, and increases the importance of correcting misclassified examples with low $p_i$. In addition, $\alpha_i$ is computed by conventional reverse frequencies, i.e., the number of total data normalized by the number of samples of the $i$-th class, and $\gamma = 2$, which is based on \citep{lin2017focal}.

All six pipelines were trained for 300 epochs, and saved for each epoch. In the binary tasks, the best model was selected by the highest TPR with over 90\% TNR. For the three-class task, the best model was selected by the highest recall of SP with over 90\% recall of UD. The optimizer was Adam \citep{kingma2014adam} with an initial learning rate of $2\times10^{-5}$. All experiments were conducted on the TensorFlow platform and performed on CyberpowerPC with single GPU (CPU: Intel Core i7-8700K@3.7GHz 6 Core, RAM:32GB \& GPU: Nvidia Geforce RTX 2080Ti).

\section{Results and Analysis}
\subsection{Experiment 1: Crack detection}
\begin{figure}
\center
\begin{subfigure}{.4\linewidth}
  \centering
  \includegraphics[width=\linewidth]{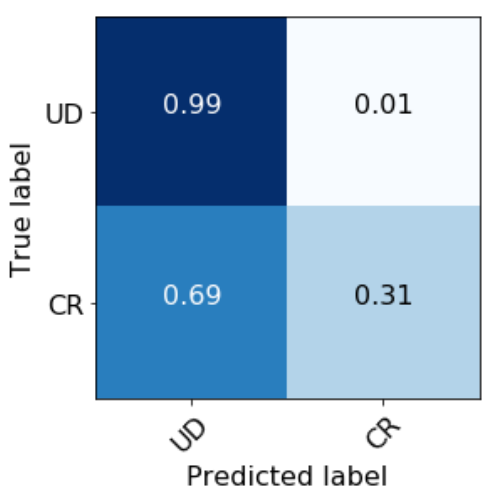}
  \caption{\label{fig_CR16_BSL}BSL}
\end{subfigure}%
\begin{subfigure}{.4\linewidth}
  \centering
  \includegraphics[width=\linewidth]{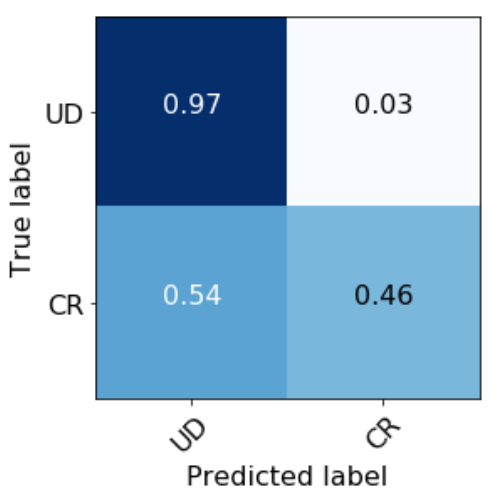}
  \caption{\label{fig_CR16_BUS}BUS}
\end{subfigure}%

\begin{subfigure}{.4\linewidth}
  \centering
  \includegraphics[width=\linewidth]{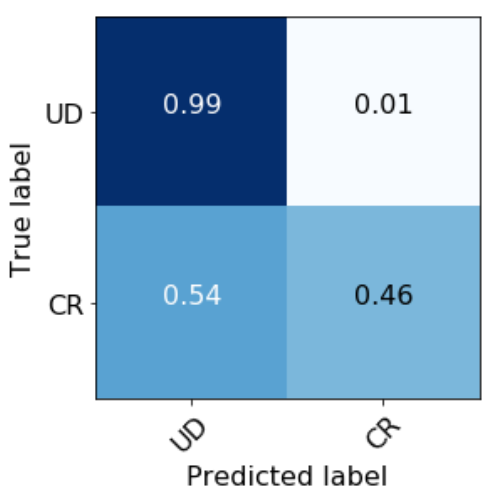}
  \caption{\label{fig_CR16_BOS_DA}BOS-DA}
\end{subfigure}
\begin{subfigure}{.4\linewidth}
  \centering
  \includegraphics[width=\linewidth]{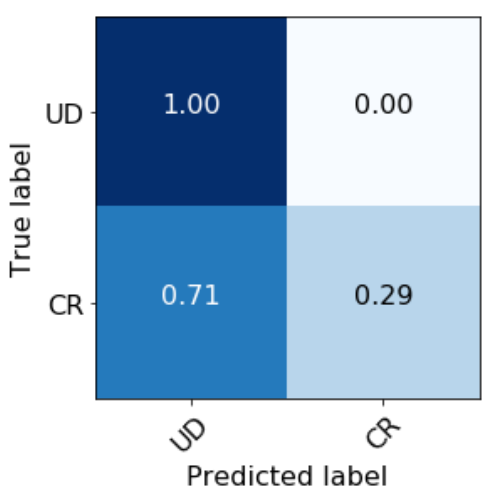}
  \caption{\label{fig_CR16_BOS_GAN}BOS-GAN}
\end{subfigure}%

\begin{subfigure}{.4\linewidth}
  \centering
  \includegraphics[width=\linewidth]{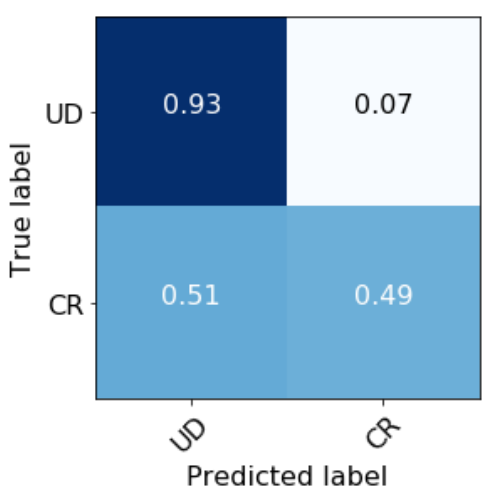}
  \caption{\label{fig_CR16_SDF}BSL-SDF}
\end{subfigure}%
\begin{subfigure}{.4\linewidth}
  \centering
  \includegraphics[width=\linewidth]{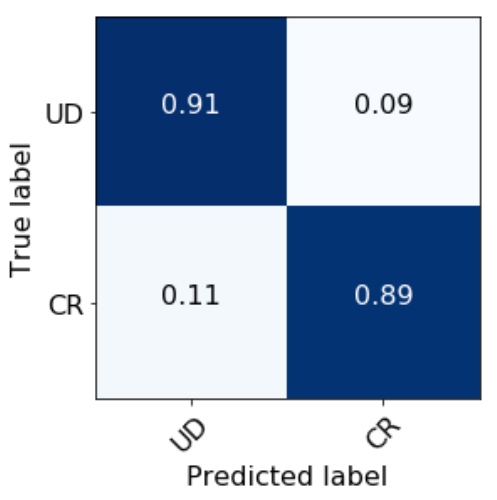}
  \caption{\label{fig_CR16_BSSGAN}BSS-GAN}
\end{subfigure}
\caption{\label{fig_CR16_CM}\enspace Normalized CM in crack detection}
\end{figure}

From Table \ref{tab_CR16_performance}, the accuracy values for all six pipelines are higher than 90\%, which are deceivingly promising. However, simply predicting all images as UD can easily lead to a 94.1 \% overall accuracy under the 16:1 class ratio. More focus should be placed on the TPR and TNR, which represent the accuracy of detecting CR and UD, respectively. The resulting TPR and TNR are shown in Table \ref{tab_CR16_performance} and Figure \ref{fig_CR16_CM}.

Starting from the low TPR of the BSL, it can be inferred that a shallow DL model can easily become biased due to extreme class imbalance (16:1). The BUS's under-sampling worked to some degree, as it improved the TPR from 31\% to 46\% without compromising the TNR too much. Similarly, BOS-DA's over-sampling helped increase the TPR to 45.7\% without too much drop in the TNR, yet its TPR (along with that of the BUS) is unsatisfactory. BOS-GAN had the worst TPR (lower than the BSL), which conforms with the observations in \citep{gao2019deep} of the low performance of directly mixing synthetic data to the pipeline. BOS-GAN is extremely biased, and it is more prone to mis-predicting data as UD, causing a meaninglessly high TNR. Three factors lead to this poor and biased behavior: (i) the risk of introducing extra parameters mentioned in Section 3.2, (ii) manually selecting augmented images is subjective, and (iii) some GAN-generated images may be ``adversarial'' images \citep{goodfellow2014explaining}. For factor (iii), although the GAN-generated images might be realistic-looking to human eyes, small feature perturbations undetectable by humans within these images might cause the classifier to make false predictions. Lastly, for the BSL-SDF pipeline, it obtained a similar performance to BUS and BOS-DA with a slight sacrifice in TNR to make up for the 3\% enhancement in TPR. BSL-SDF is still biased, for which we can infer that even though SDF improved model initialization, it did not help with solving the class imbalance issue.

In general, the five pipelines above are unsatisfactory in crack detection under a UD:CR = 16:1 class ratio. These five pipelines have TPR below 50\% and misleadingly high TNR, implying severe biases towards the UD class. Accordingly, these pipelines can only detect less than 50\% of all cracked structures or components, which is unacceptable in practice. On the contrary, the BSS-GAN model not only maintained an equally good TNR (over 92\%) as others, but its TPR was also substantially higher (about 90\%), indicating a nearly unbiased performance. Moreover, BSS-GAN is efficient in training, and it has an advantage over other models in the following sense: compared with BUS, BSS-GAN can utilize all accessible data, which provide additional information; compared with BOS-DA, BOS-GAN, and BSL-SDF, the DA process is hidden and involved in the training process, with no extra data storage or multi-step training required.

\begin{figure}
\centering
\includegraphics[width=0.75\linewidth]{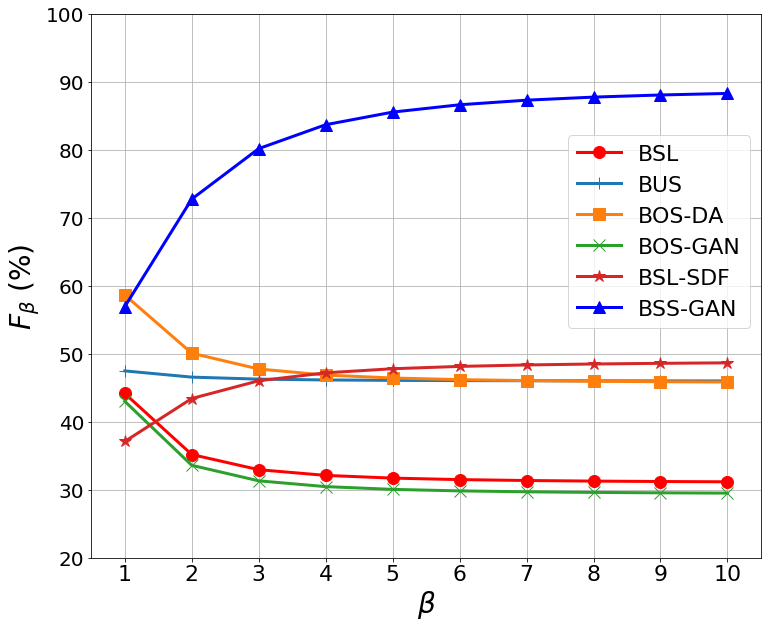}
\caption{\enspace $F_{\beta}$ with varying $\beta$ in crack detection\label{fig_CR_16_F_Beta}}
\end{figure}

$F_2$ \& $F_5$ scores were computed to take the recall (TPR) and precision into account. As aforementioned, the selection of $\beta$ usually depends on its real-world interpretation. To avoid loss of generality, $F_\beta$ scores with varying $\beta$ values are plotted in Figure \ref{fig_CR_16_F_Beta}. It is observed that starting from $\beta=1$ (weighting recall and precision equally), BSS-GAN and BSL-SDF have increasing trends while BSL, BOS-DA, and BOS-GAN show decreasing trends. In addition, the $F_\beta$ values converge to recall scores (TPR) as $\beta$ increases. In SHM, it is more crucial to reduce FN than FP, so a large $\beta$ is preferred. From Figure \ref{fig_CR_16_F_Beta}, as $\beta$ becomes larger, $F_\beta$ from BSS-GAN exceeds those of other models with growing differences, suggesting its superiority in crack detection problems with high class imbalance. 

\begin{table}
\centering
\caption{\enspace Classification performance in CR detection ($\%$)}
\label{tab_CR16_performance}
\vspace*{-1mm}
\begin{tabular}{c|c|c|c|c|c}
\hline
\textbf{Pipeline} & \textbf{TPR} & \textbf{TNR} & $\mathbf{F_2}$ & $\mathbf{F_5}$ & \textbf{Accuracy} \\ \hline
BSL            & 31.0   & 99.4   & 35.2   & 31.7 & 95.4      \\ \hline
BUS            & 46.0   & 97.0   & 46.6   & 46.1 & 94.0      \\ \hline
BOS-DA         & 45.7   & 99.4   & 50.1   & 46.5 & 96.2      \\ \hline
BOS-GAN        & 29.3   & 99.5   & 33.6   & 30.1 & 95.4      \\ \hline
BSL-SDF            & 49.0   & 92.8   & 43.4   & 47.8  & 90.2      \\ \hline
BSS-GAN        & \textbf{89.3}   & 92.2   & \textbf{72.8}   & \textbf{85.6} & 92.1      \\ \hline
\end{tabular}
\end{table}

\subsection{Experiment 2: Spalling detection}

\begin{table}
\centering
\caption{\enspace Classification performance in SP detection ($\%$)}
\label{tab_SP32_performance}
\vspace*{-1mm}
\begin{tabular}{c|c|c|c|c|c}
\hline
\textbf{Pipeline} & \textbf{TPR} & \textbf{TNR} & $\mathbf{F_2}$ & $\mathbf{F_5}$ & \textbf{Accuracy} \\ \hline
BSL            & 62.0   & 99.8   & 66.1   &   62.8  & 98.6      \\ \hline
BUS            & 84.7   & 97.1   & 73.2   &   82.2  & 96.7      \\ \hline
BOS-DA         & 83.3   & 99.9   & 85.5  & 83.7  & 99.4      \\ \hline
BOS-GAN        & 64.0   & 99.9   & 68.6   &   64.8  & 99.0      \\ \hline
BSL-SDF            & 84.0   & 92.8   & 62.3   &   78.7  & 93.8      \\ \hline
BSS-GAN        & \textbf{98.0}   & 95.8   &   77.6  & \textbf{93.3}  & 95.9     \\ \hline
\end{tabular}
\end{table}

\begin{figure}
\center
\begin{subfigure}{.4\linewidth}
  \centering
  \includegraphics[width=\linewidth]{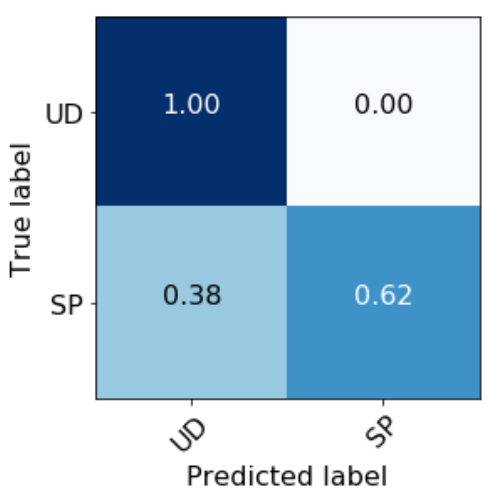}
  \caption{\label{fig_SP32_BSL}BSL}
\end{subfigure}
\begin{subfigure}{.4\linewidth}
  \centering
  \includegraphics[width=\linewidth]{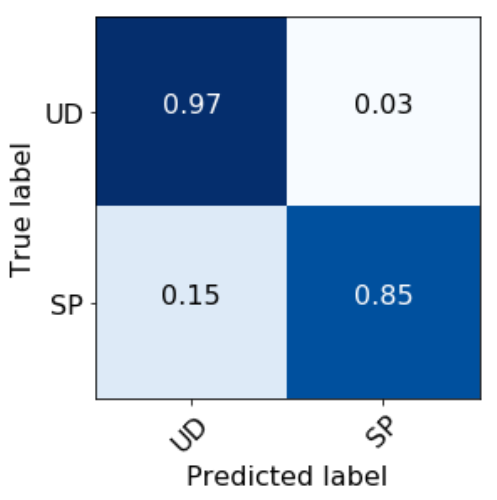}
  \caption{\label{fig_SP32_BUS}BUS}
\end{subfigure}%

\begin{subfigure}{.4\linewidth}
  \centering
  \includegraphics[width=\linewidth]{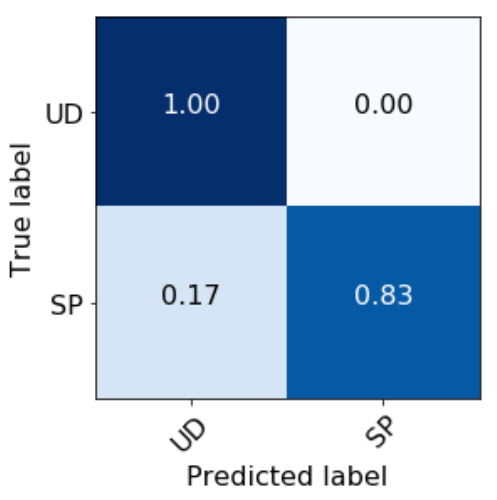}
  \caption{\label{fig_SP32_BOS_DA}BOS-DA}
\end{subfigure}
\begin{subfigure}{.4\linewidth}
  \centering
  \includegraphics[width=\linewidth]{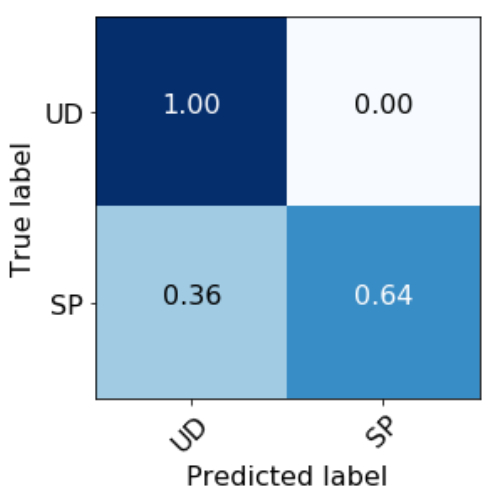}
  \caption{\label{fig_SP32_BOS_GAN}BOS-GAN}
\end{subfigure}

\begin{subfigure}{.4\linewidth}
  \centering
  \includegraphics[width=\linewidth]{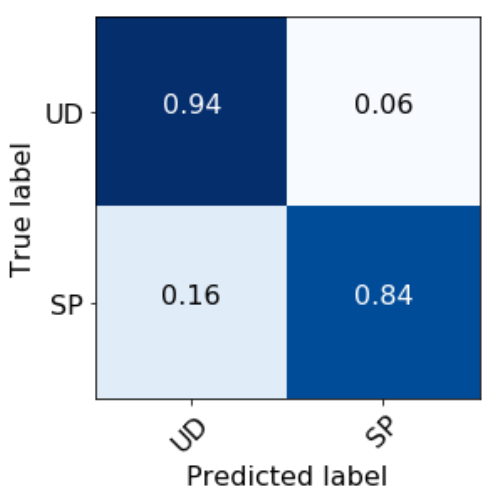}
  \caption{\label{fig_SP32_SDF}BSL-SDF}
\end{subfigure}%
\begin{subfigure}{.4\linewidth}
  \centering
  \includegraphics[width=\linewidth]{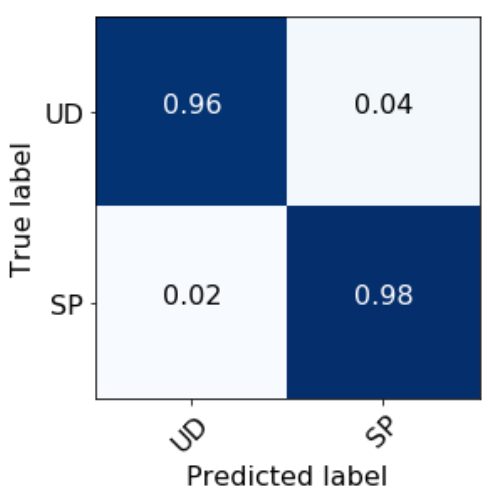}
  \caption{\label{fig_SP32_BSSGAN}BSS-GAN}
\end{subfigure}
\caption{\label{fig_SP32_CM}\enspace Normalized CM in spalling detection}
\end{figure}

Even though the imbalanced UD:SP = 32:1 class ratio in this experiment is twice that of the crack detection task, from Table \ref{tab_SP32_performance} and Figure \ref{fig_SP32_CM}, the performance of all pipelines are better than experiment 1. This observation can be partially explained by different degrees of visual pattern similarity among UD, CR and SP. CR images (Figure \ref{fig_samples_b}) are similar to UD images (Figure \ref{fig_samples_a}) except for the appearance of surface fissures or fine cracks. On the contrary, SP images (Figure \ref{fig_samples_c}) are more dissimilar, where the areas of spalling break the surface patterns in both color and texture, making the SP features more distinguishable. In this experiment, all pipelines obtained satisfactory TNR, and over 50\% TPR values. As in experiment 1, BSL and BOS-GAN had the lowest TPR, which again showed the ineffectiveness of directly mixing synthetic data to the pipeline. BUS, BOS-DA, and BSL-SDF achieved similar performance, with TPR values reaching nearly 84\%. BSS-GAN achieved the highest TPR (consistent with its crack detection performance in experiment 1). It not only maintained a high TNR (95.8\%) as with other pipelines, but also improved the TPR from BSL's 62.0\% to a surprising 98.0\%, which is nearly 14\% higher than those of BUS, BOS-DA, and BSL-SDF. Clearly, the BSS-GAN outperformed the other five pipelines.

\begin{figure}
\centering
\includegraphics[width=0.75\linewidth]{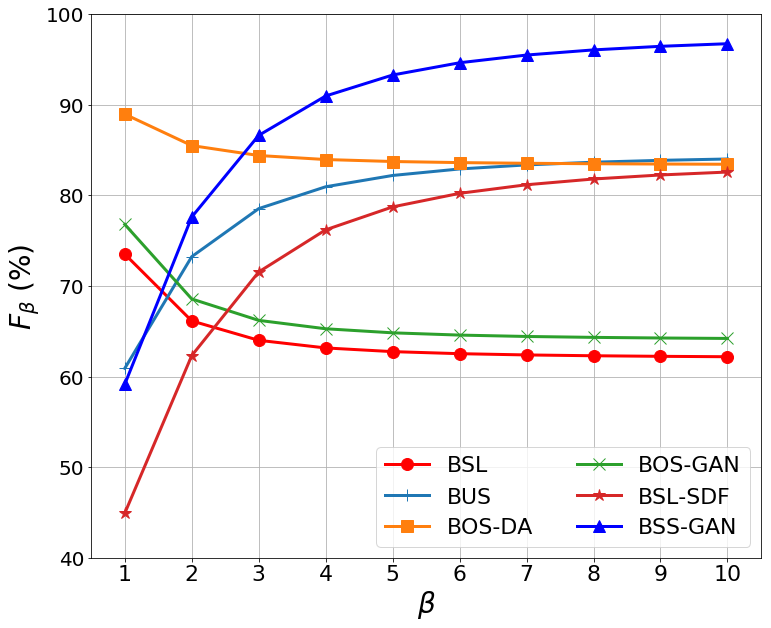}
\caption{\enspace $F_{\beta}$ with varying $\beta$ in spalling detection\label{fig_SP_32_F_Beta}}
\end{figure}

In spalling detection, due to high class imbalance, a small decrease in TNR will over-emphasize the increase of FP, leading to a higher $F_2$ score. For example, a mere 4.1\% drop in TNR from BOS-DA to BSS-GAN makes $F_2$ of BOS-DA higher, although the TPR of BOS-DA is 15\% lower than that of BSS-GAN. Under the UD:SP = 32:1 class ratio, the $F_2$ score does not place enough emphasis on the recall (TPR). Thus, a larger $\beta = 5$ is more informative. According to Figure \ref{fig_SP_32_F_Beta}, both BSL and BOS-GAN share similar values and trends, while BUS, BOS-DA, and BSL-SDF converge to the same value after $\beta=5$. As $\beta$ increases, especially beyond $\beta=5$, BSS-GAN significantly outperforms other pipelines. 

\subsection{Experiment 3: Damage pattern recognition}
\begin{figure}
\center
\begin{subfigure}{.4\linewidth}
  \centering
  \includegraphics[width=\linewidth]{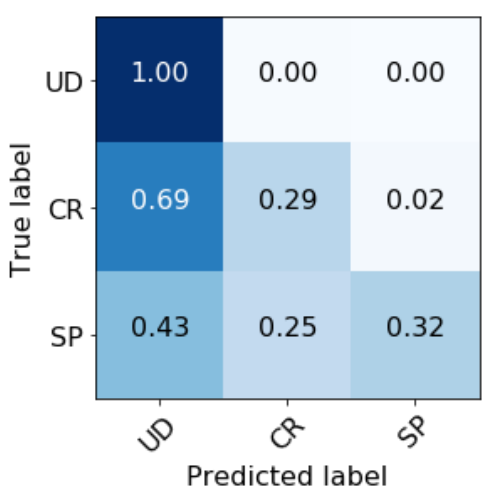}
  \caption{\label{fig_TRI_BSL}BSL}
\end{subfigure}
\begin{subfigure}{.4\linewidth}
  \centering
  \includegraphics[width=\linewidth]{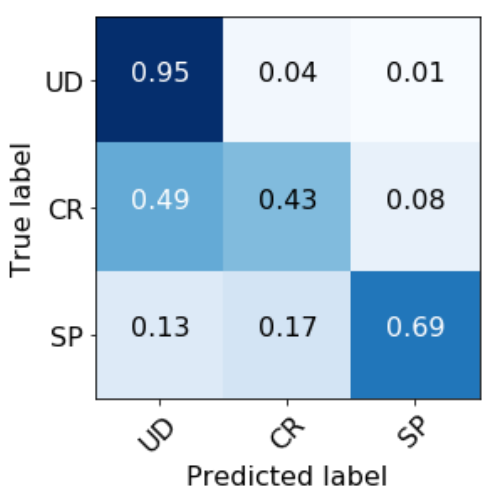}
  \caption{\label{fig_TRI_BUS}BUS}
\end{subfigure}%

\begin{subfigure}{.4\linewidth}
  \centering
  \includegraphics[width=\linewidth]{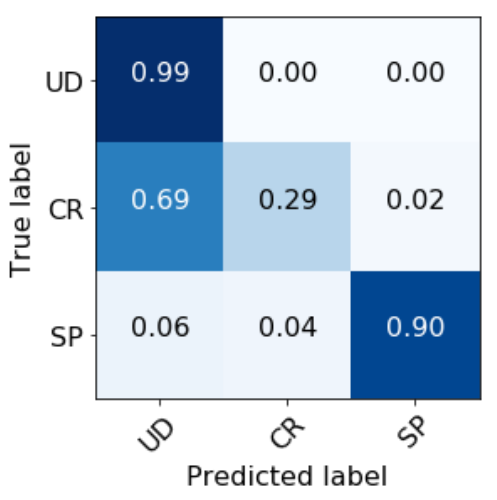}
  \caption{\label{fig_TRI_BOS_DA}BOS-DA}
\end{subfigure}
\begin{subfigure}{.4\linewidth}
  \centering
  \includegraphics[width=\linewidth]{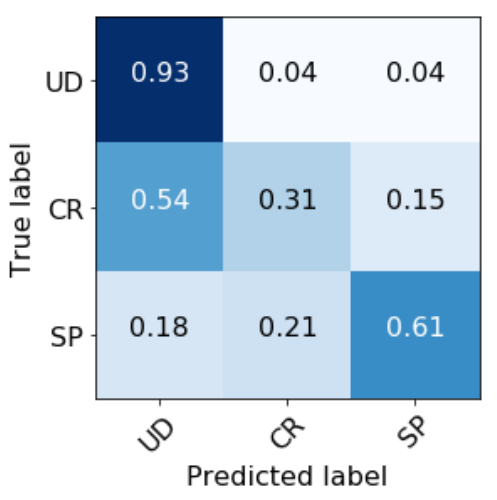}
  \caption{\label{fig_TRI_BOS_GAN}BOS-GAN}
\end{subfigure}

\begin{subfigure}{.4\linewidth}
  \centering
  \includegraphics[width=\linewidth]{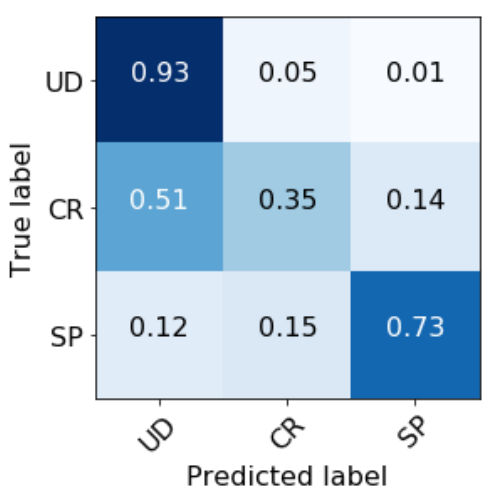}
  \caption{\label{fig_TRI_SDF}BSL-SDF}
\end{subfigure}%
\begin{subfigure}{.4\linewidth}
  \centering
  \includegraphics[width=\linewidth]{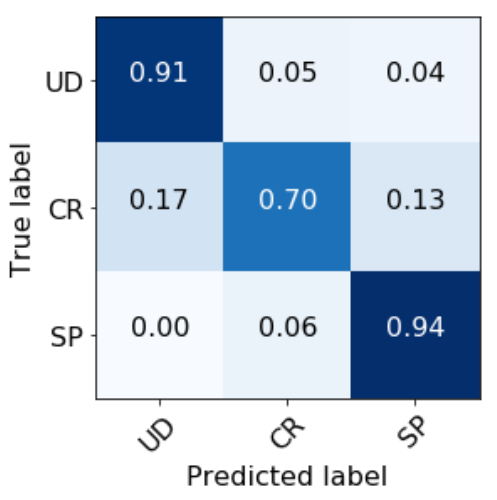}
  \caption{\label{fig_TRI_BSSGAN}BSS-GAN}
\end{subfigure}
\caption{\label{fig_TRI_CM}\enspace Normalized CM in damage pattern recognition}
\end{figure}

In this experiment, a more complex multi-class classification was investigated, where the imbalanced ratio for UD:CR:SP is 32:2:1. The performance results are listed in Table \ref{tab_TRI_performance} and Figure \ref{fig_TRI_CM}. As with the two binary experiments 1 \& 2, the BSL was biased in favor of UD with low recall values (around 30\%) for both minority classes CR \& SP. BOS-GAN was the second worst pipeline in terms of class recall (31.3\% \& 60.7\% for CR \& SP, respectively). BUS, BOS-DA and BSL-SDF did not perform well either, as their improvements in SP recall were merely sacrifices of the CR recall. For example, BOS-DA's SP recall reached 90\%, but its CR recall remained low at 29\%.

In general, the first five pipelines share a common drawback, namely the bias against CR, as characterized by their low CR recall values. This issue is attributed to the high visual similarity between UD and CR. GAN-based DA worsens this issue by generating images with blended features between UD and CR. On the contrary, the BSS-GAN pipeline outperformed others with 90\% UD recall, 70\% CR recall, and 94\% SP recall. Additionally, the BSS-GAN only misclassified 6\% of the SP images as CR, and no SP images were misidentified as UD. It is thus much more reliable in detecting severer damage.

\begin{table}[ht]
\centering
\caption{\enspace Classification performance in damage pattern recognition ($\%$)}
\label{tab_TRI_performance}
\vspace*{-1mm}
\begin{tabular}{c|c|c|c|c}
\hline
\multirow{2}{*}{\textbf{Pipeline}} & \multirow{2}{*}{\textbf{Accuracy}} & \multicolumn{3}{c}{\textbf{Recall}} \\ \cline{3-5}
& & \textbf{UD} & \textbf{CR} & \textbf{SP} \\ \hline
BSL            & 93.6   & 99.5   & 28.7   & 32.0  \\ \hline
BUS            & 91.7   & 95.4   & 42.6   & 69.3  \\ \hline
BOS-DA         & 95.2   & 99.5   & 29.0   & 90.0  \\ \hline
BOS-GAN        & 88.3   & 92.7   & 31.3   & 60.7  \\ \hline
BSL-SDF            & 89.6   & 93.5   & 35.0   & 73.3  \\ \hline
BSS-GAN        & 89.7   & 90.9   & \textbf{70.0}   & \textbf{94.0}   \\ \hline
\end{tabular}
\end{table}

\subsection{Investigation of failure cases}
\begin{figure}
\centering
\begin{subfigure}{.42\linewidth}
  \centering
  \includegraphics[width=\linewidth]{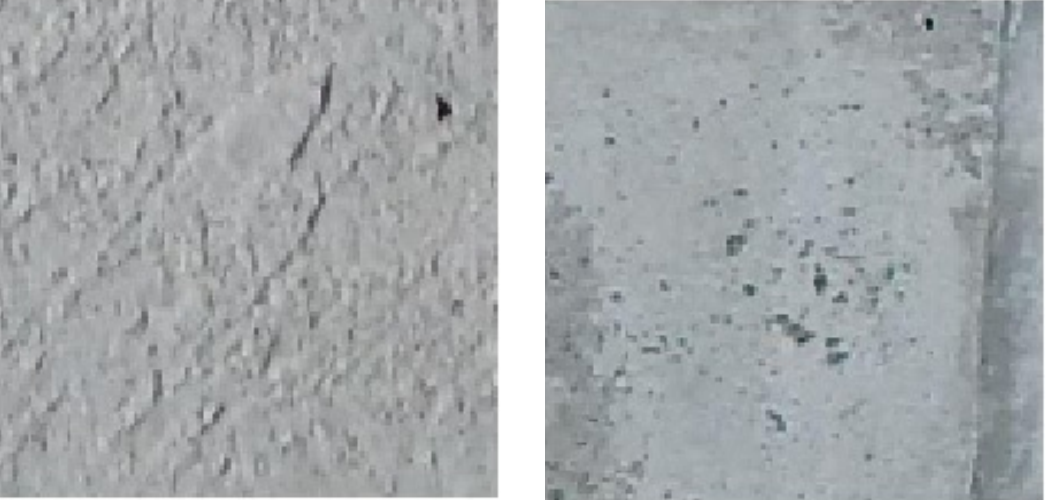}
  \caption{\label{fig_a_UD_CR}Truth:UD Prediction:CR}
\end{subfigure} \hspace{5pt}
\begin{subfigure}{.42\linewidth}
  \centering
  \includegraphics[width=\linewidth]{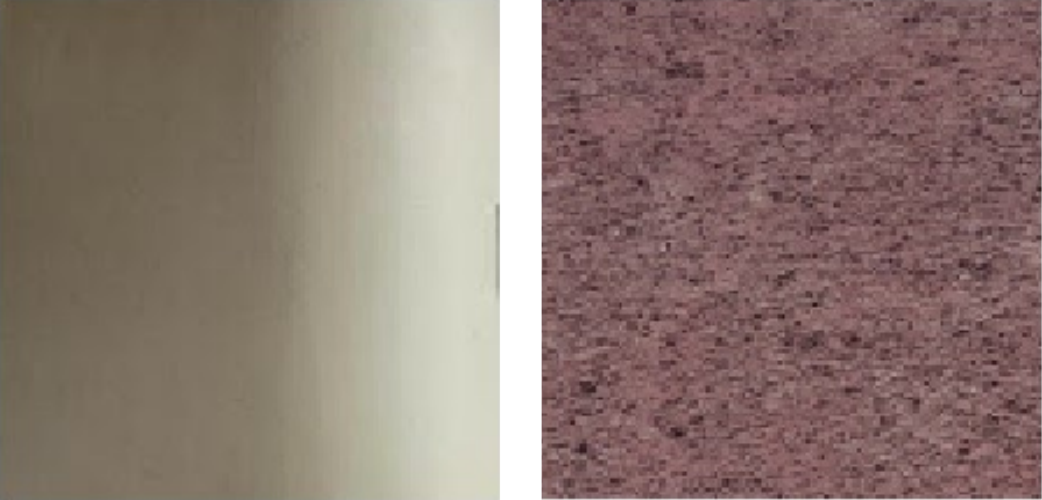}
  \caption{\label{fig_b_UD_SP}Truth:UD Prediction:SP}
\end{subfigure}

\begin{subfigure}{.42\linewidth}
  \centering
  \includegraphics[width=\linewidth]{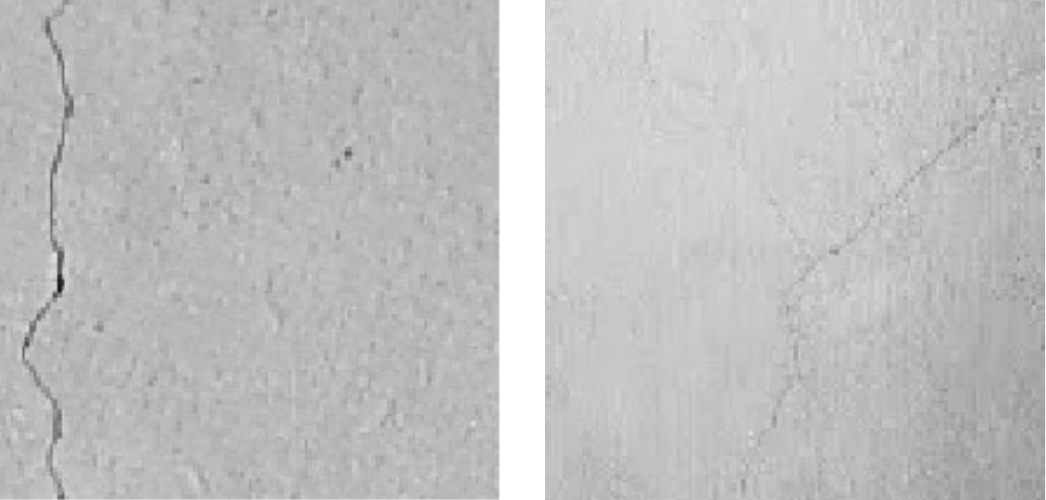}
  \caption{\label{fig_c_CR_UD}Truth:CR Prediction:UD}
\end{subfigure} \hspace{5pt}
\begin{subfigure}{.42\linewidth}
  \centering
  \includegraphics[width=\linewidth]{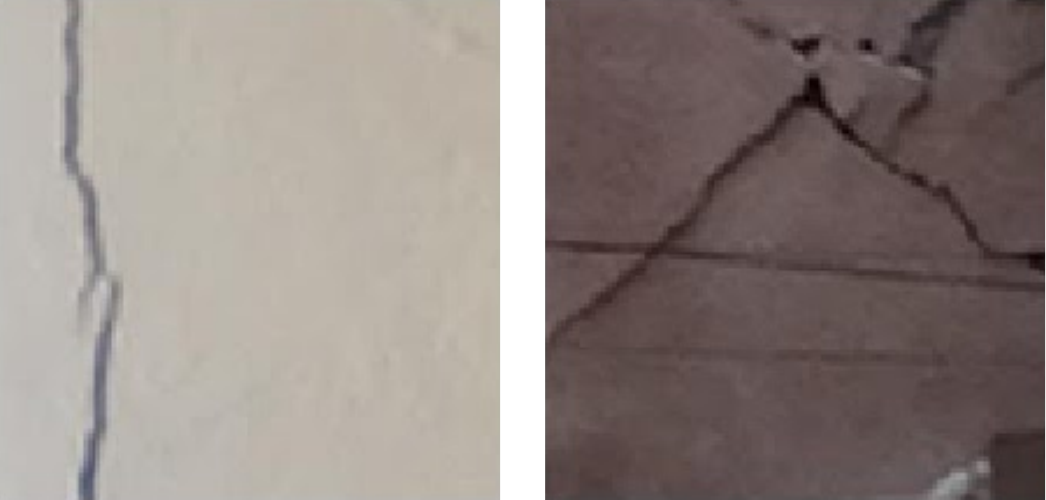}
  \caption{\label{fig_d_CR_SP}Truth:CR Prediction:SP}
\end{subfigure}

\begin{subfigure}{.42\linewidth}
  \centering
  \includegraphics[width=\linewidth]{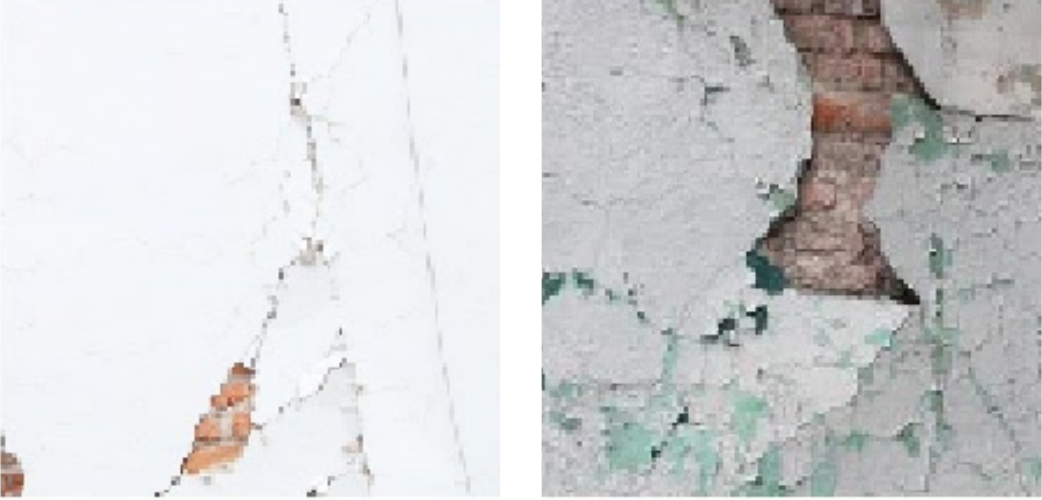}
  \caption{\label{fig_e_SP_CR}Truth:SP Prediction:CR}
\end{subfigure}
\caption{\label{fig_failure_cases}\enspace Sample images of BSS-GAN's failure cases}
\end{figure}

For more generality, the failure cases of the three-class damage pattern recognition are investigated. Selected examples of images that are misclassified are presented in Figure \ref{fig_failure_cases}.

From the CM in Figure \ref{fig_TRI_BSSGAN}, it is known that among misclassified images, the ratios of UD images misclassified as CR (5\%) or SP (4\%), and SP images misclassified as CR (6\%) are on the same order of scale (first and third row), while the ratios of CR images misclassified as UD (17\%) or SP (13\%) are larger (middle row). For misclassified UD images, wall texture and color contrast are the main contributors to the inaccuracies. For example, the model tends to mistaken rougher wall textures, Figures \ref{fig_a_UD_CR} \& \ref{fig_b_UD_SP}, as damaged (SP or CR), and color contrasts of large patches, Figure \ref{fig_b_UD_SP}, as SP. For misclassified SP images, no images are classified as UD, but some images of mixed damage types (with both CR and SP features), Figure \ref{fig_e_SP_CR}, are misjudged by the model. It is noted that images of mixed damage types are usually labeled as SP).

The CR images are more likely to be misclassfied than UD or SP images, Figures \ref{fig_c_CR_UD} \& \ref{fig_d_CR_SP}. These CR images contain features in between UD and SP. Accordingly, the CR samples lie at the intersection of UD and SP in the feature manifold and are particularly hard to be distinguished by the model. Crack sizes (widths) directly affect the model's judgement on whether an image is more likely to be misclassified as UD or SP.

\subsection{Investigation of synthetic image quality}
\begin{figure}
\centering
\begin{subfigure}{.4\linewidth}
  \centering
  \includegraphics[width=\linewidth]{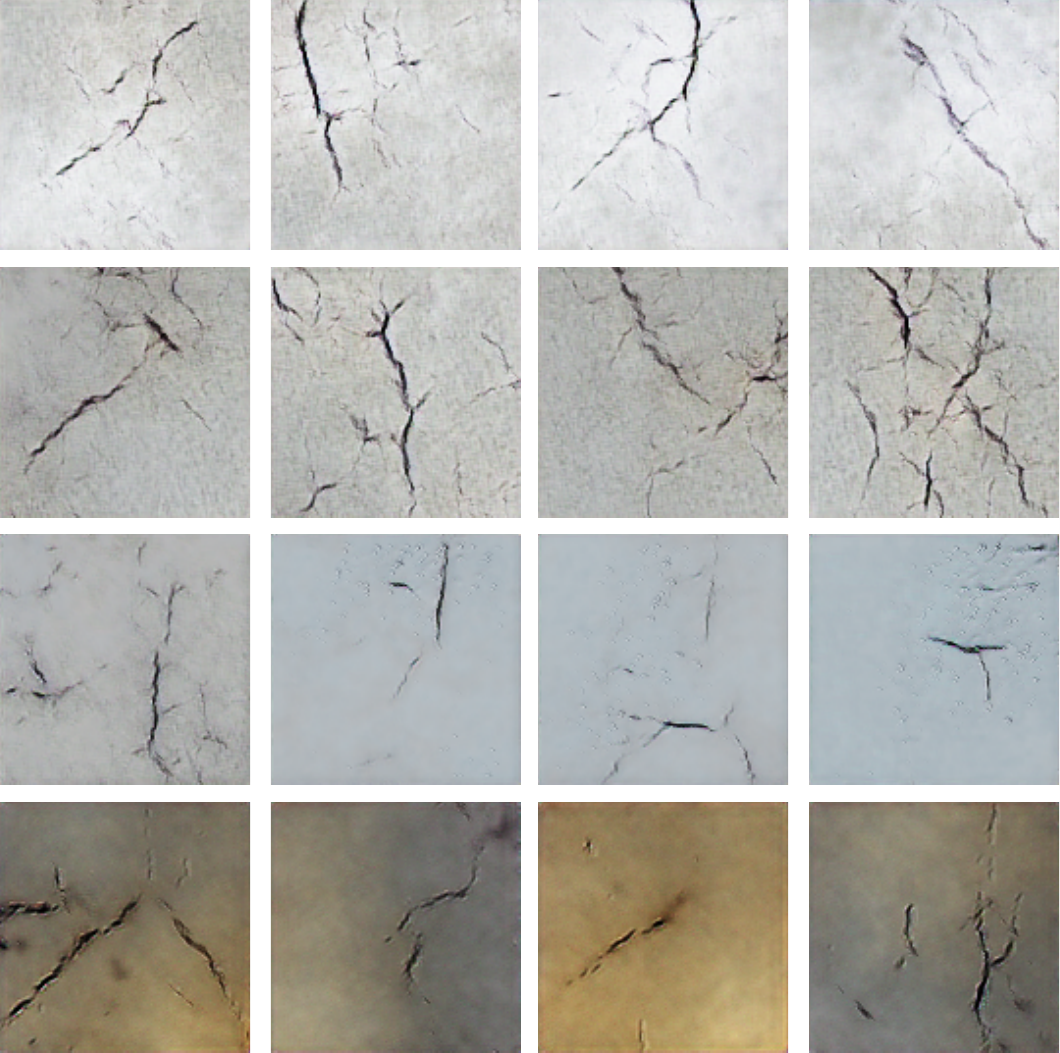}
  \caption{\label{fig_fake_CR_a}Ordinary GAN}
\end{subfigure}%
\hspace{3mm}
\begin{subfigure}{.4\linewidth}
  \centering
  \includegraphics[width=\linewidth]{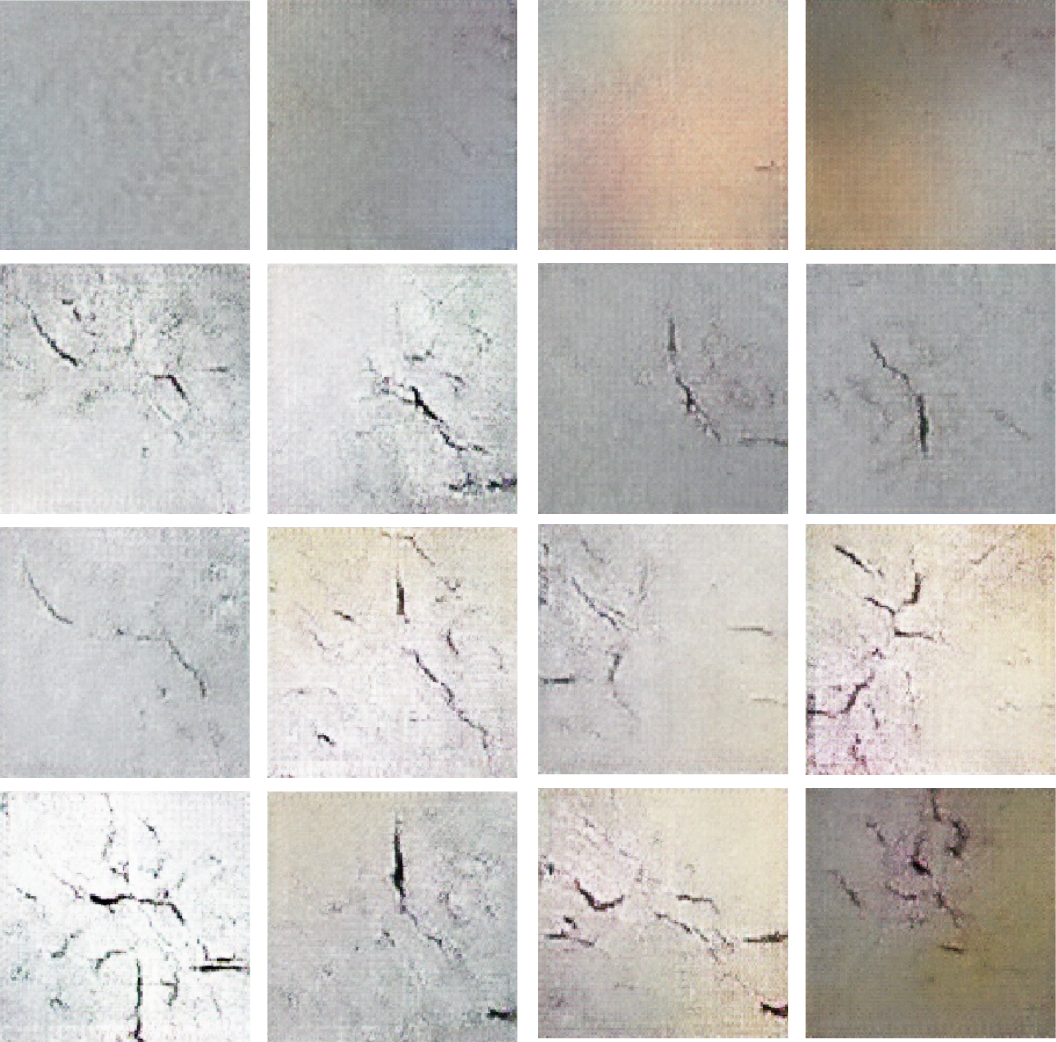}
  \caption{\label{fig_fake_CR_b}BSS-GAN}
\end{subfigure}%
\caption{\label{fig_fake_images_CR}\enspace Sample synthetic images in crack detection}
\end{figure}

\begin{figure}
\centering
\begin{subfigure}{.4\linewidth}
  \centering
  \includegraphics[width=\linewidth]{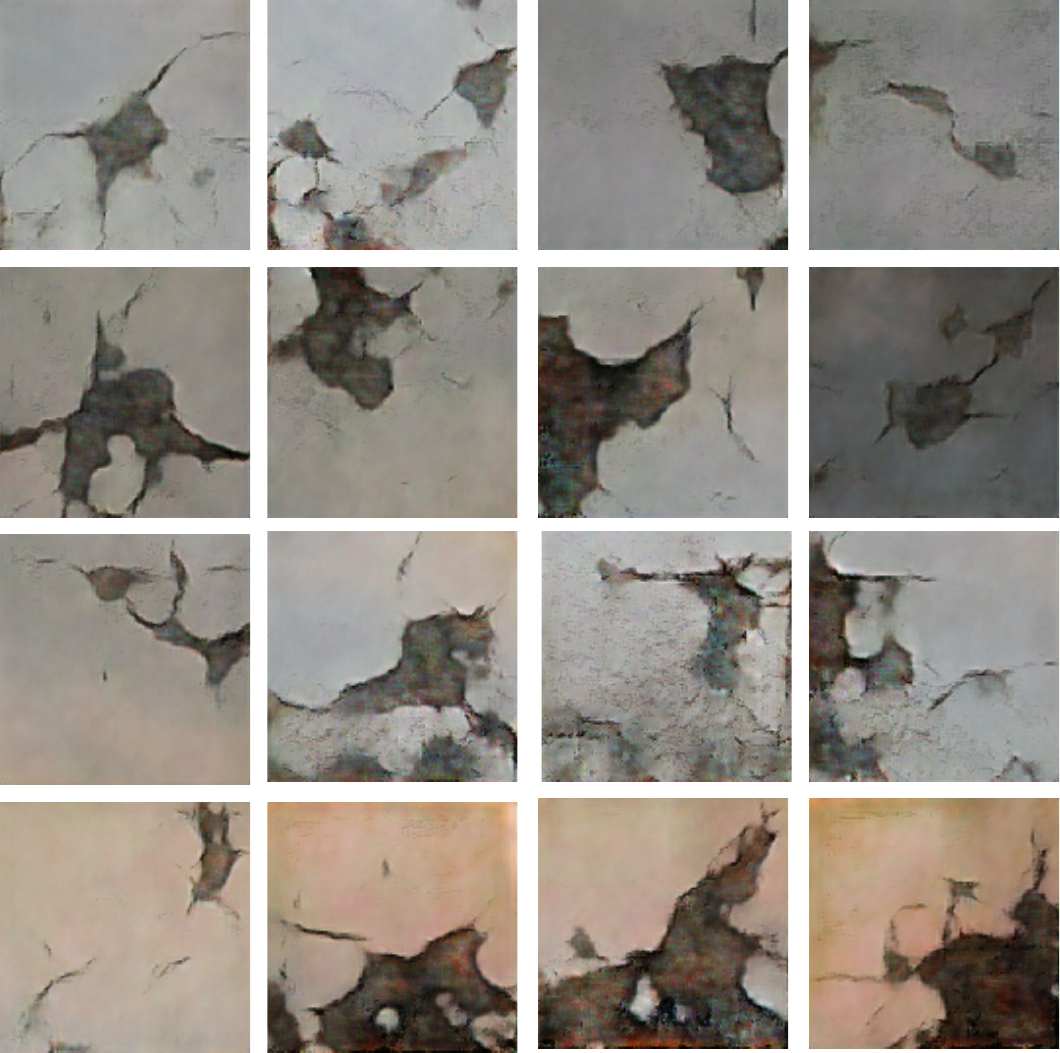}
  \caption{\label{fig_fake_SP_a}Ordinary GAN}
\end{subfigure}%
\hspace{3mm}
\begin{subfigure}{.4\linewidth}
  \centering
  \includegraphics[width=\linewidth]{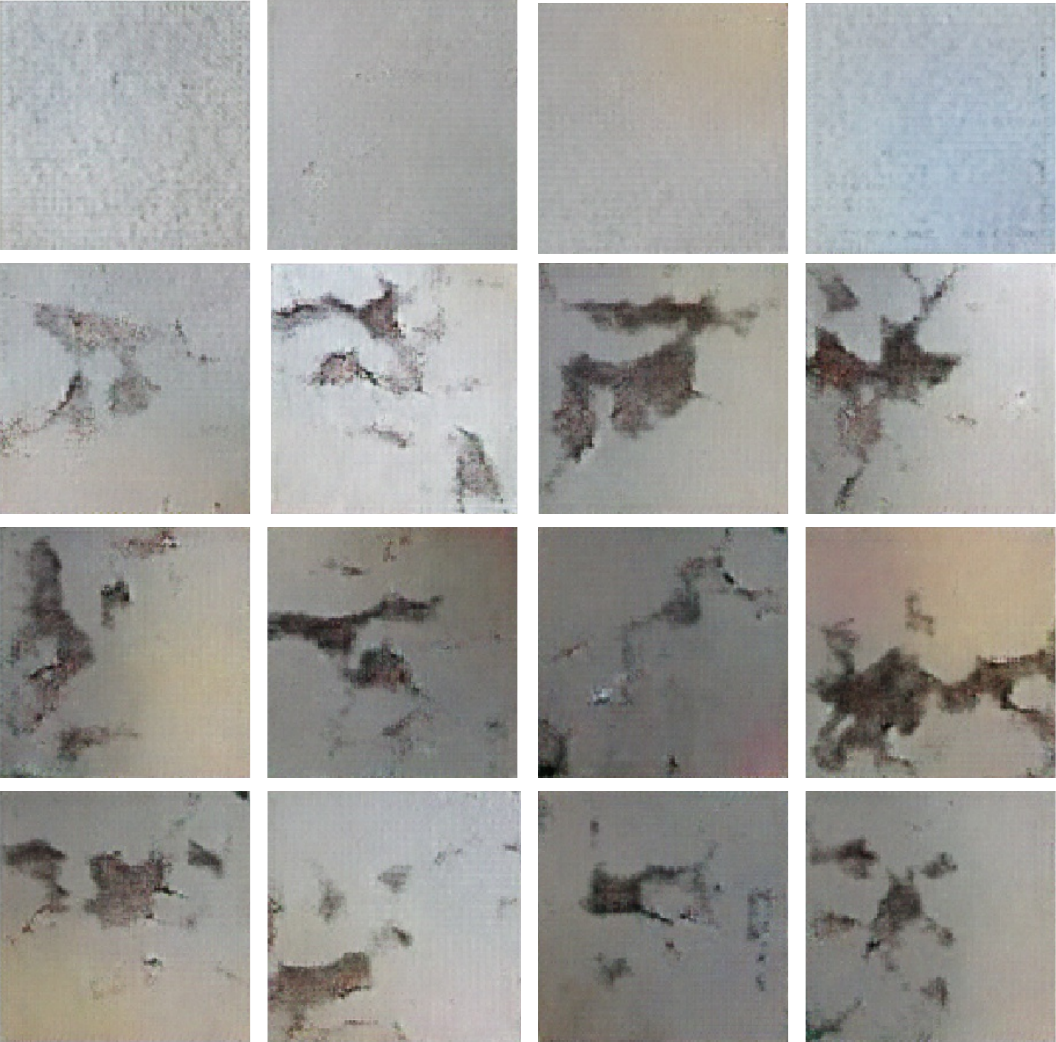}
  \caption{\label{fig_fake_SP_b}BSS-GAN}
\end{subfigure}%
\caption{\label{fig_fake_images_SP}\enspace Sample synthetic images in spalling detection}
\end{figure}

\begin{figure}
\centering
\includegraphics[width=\linewidth]{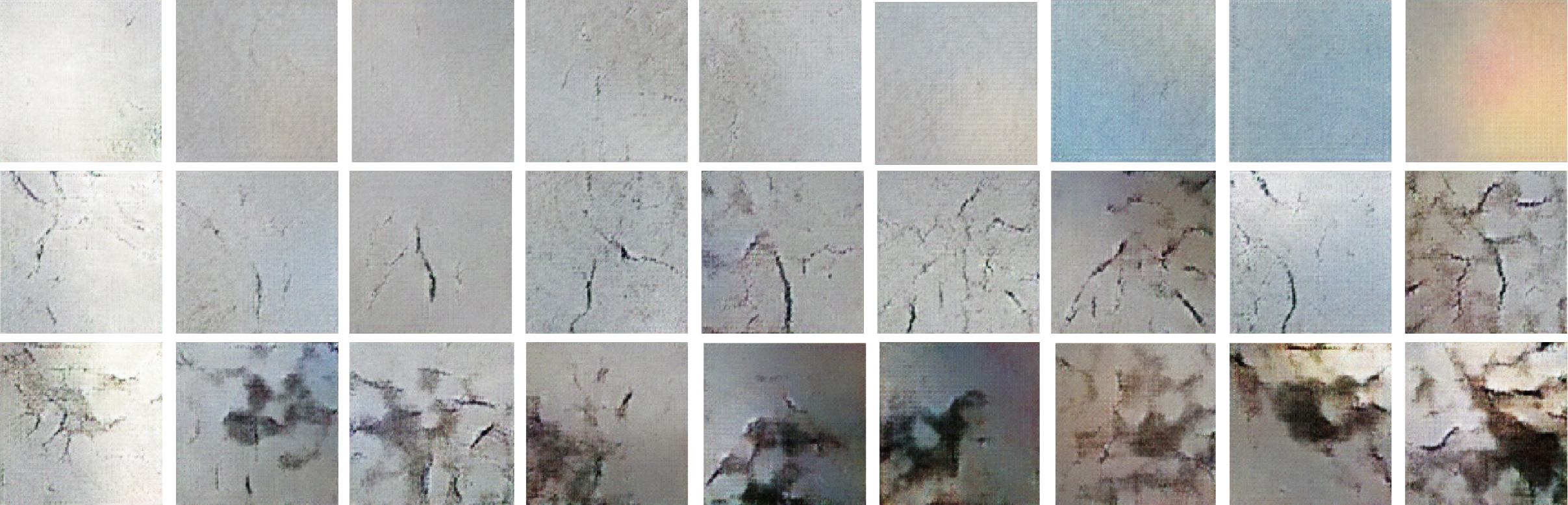}
\caption{\label{fig_fake_images_TRI}\enspace Sample synthetic images in damage pattern recognition generated by BSS-GAN}
\end{figure}

In this section, synthetic images generated from well-trained BSS-GAN models in the above experiments were compared with those generated by ordinary GAN in BOS-GAN and BSL-SDF pipelines, Figures \ref{fig_fake_images_CR}, \ref{fig_fake_images_SP} \& \ref{fig_fake_images_TRI}. It is noted that the generator in the BSS-GAN was trained using mixed-class images instead of class-specific (minority class) as in BOS-GAN and BSL-SDF, so it learned a mixed distribution of UD, CR \& SP. For example, for crack detection, BSS-GAN was capable to generate both synthetic UD and CR images, Figure \ref{fig_fake_CR_b}. 

Overall, there is no obvious mode collapse issue (the generator only produces limited varieties of images) in either the ordinary GAN or BSS-GAN. Besides basic visual features like textures and colors, the generator in both models can generate images with a variety of more complex features, e.g., crack orientation, location and width, and spalling shape, location and area. As mentioned in \citep{gao2019deep}, structural images have complex and mixed distributions, which make it difficult for GAN to generate clear and class-discriminative images. However, conditioning operation (considering class information related to a specific class of images) makes the ordinary GAN capable of generating higher quality images towards that class. For example, when training the GANs of BOS-GAN and BSL-SDF pipelines for crack detection, only feeding minority-class (CR) images can be viewed as one type of conditioning operation, which significantly reduces the data distribution complexity. This cleaner and easier sub-distribution boosts the learning process, and avoids the model being trapped in certain local saddle points leading to mode collapse. Thus, in Figure \ref{fig_fake_CR_a}, the synthetic images show very realistic visual qualities along with variety. On the other hand, the generator in the BSS-GAN was trained with all images in an unsupervised manner. Thus, it had to learn a mixed distribution from both UD and CR. As a result, the synthetic images generated by BSS-GAN have features of UD, CR, or even the intermediate (mixed) state. The ordinary GAN using all images without any conditioning operations was unable to generate reasonable structural images and occasionally led to mode collapse issues similar to the failure cases in \citep{gao2019deep}. However, the balanced-batch sampling mechanism strictly enforces equal occurrences of each class within one training batch, which make the model less biased to certain classes and thus relieve the mode collapse issue. To show this, in Figure \ref{fig_fake_CR_b}, synthetic images in the first row are smooth and resemble UD, while the remaining images resemble CR, but are somewhat blurry.

Another possible explanation of the differences in image quality is the loss function. Unlike ordinary GAN, the loss function in BSS-GAN focuses more on classification than the quality of generated images. This is reflected by the supervised cross-entropy loss (Equation \ref{eq_supervised_loss}). On the contrary, the ordinary GAN only utilizes the unsupervised loss  (Equation \ref{eq_unsupervised_loss}), which is more about feature learning than classification. These different training objectives influence the performance of the generator even though the same network architecture was used. It is thus inferred that BSS-GAN trades off its generator performance for more improvement in the discriminator's classification capabilities.

In summary, the ordinary GAN with class conditioning is able to generate higher-quality images than BSS-GAN by human judgement under the 128$\times$128 pixel resolution. However, according to \citep{salimans2016improved, gao2019deep}, sometimes the realistic-looking images may not align well with the correct features for the classification, e.g., an image is recognized as UD by the model, but it looks like CR by a human. Thus, if the synthetic images selection is through human interaction, using these synthetic data may worsen the classifier's performance as shown in BOS-GAN \& BSL-SDF. On the contrary, for BSS-GAN, the steps from synthetic image generation to feature learning and classification are automatically and implicitly embedded in the training process, which is characterized by the game-theoretic competition between $D$ \& $G$, and no human interaction is required. These characteristics help the BSS-GAN learn meaningful representations better and improve both training efficiency and the discriminative performance, which are supported by the results in Figures \ref{fig_CR16_BSSGAN}, \ref{fig_SP32_BSSGAN} \& \ref{fig_TRI_BSSGAN}.


\subsection{Experiment 4: Unsupervised feature learning}
\begin{figure}
\center
\begin{subfigure}{.4\linewidth}
  \centering
  \includegraphics[width=\linewidth]{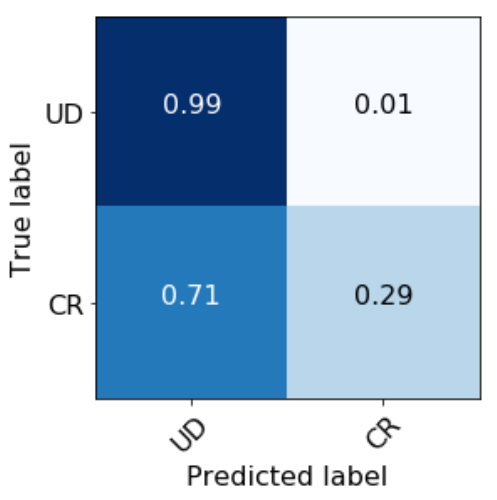}
  \caption{\label{fig_Unsup_BSL}BSL}
\end{subfigure}%
\begin{subfigure}{.4\linewidth}
  \centering
  \includegraphics[width=\linewidth]{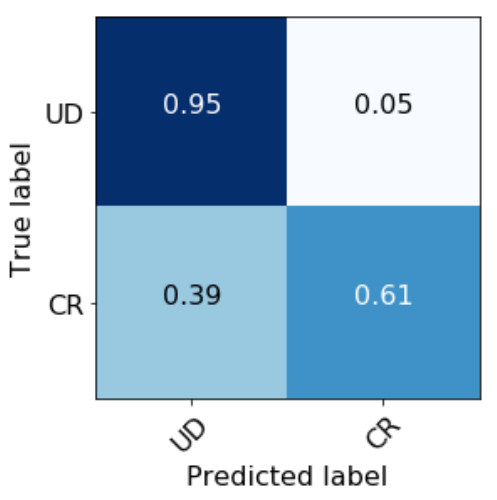}
  \caption{\label{fig_Unsup_BSSGAN_0}BSS-GAN-0}
\end{subfigure}%

\begin{subfigure}{.4\linewidth}
  \centering
  \includegraphics[width=\linewidth]{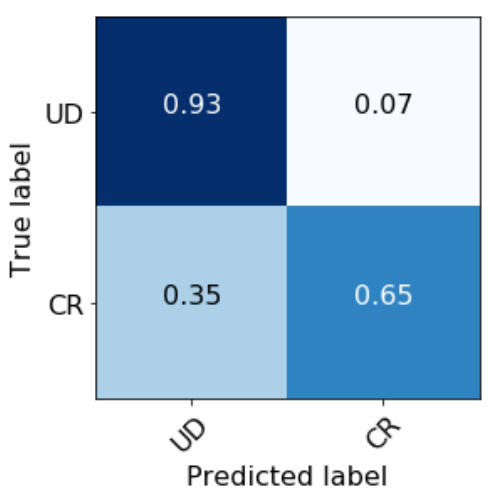}
  \caption{\label{fig_Unsup_BSSGAN_40}BSS-GAN-50}
\end{subfigure}
\begin{subfigure}{.4\linewidth}
  \centering
  \includegraphics[width=\linewidth]{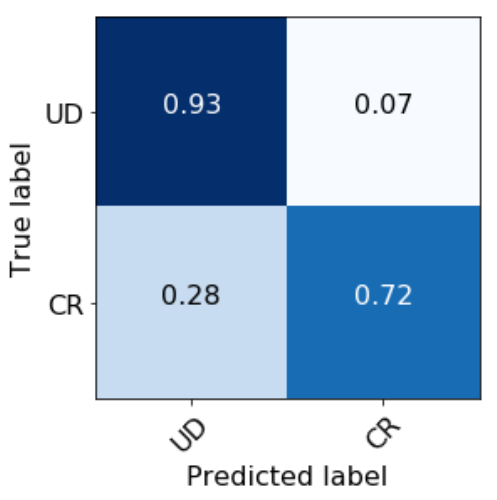}
  \caption{BSS-GAN-100}
  \label{fig_Unsup_BSSGAN_80}
\end{subfigure}%

\caption{\label{fig_Unsup}\enspace Normalized CM of four cases under a reduced-scale dataset with additional unlabeled data}
\end{figure}

In this part, both BSL and BSS-GAN were initially trained using only 2,040 labeled samples (20\% of the training data). More unlabeled data were progressively added to subsequent BSS-GAN trials (0\% for BSS-GAN-0, 50\% for BSS-GAN-50 \& 100\% for BSS-GAN-100 of the remaining 8,160 samples). Results of the four cases are shown in Table \ref{tab_Unsup} and Figure \ref{fig_Unsup}. 

According to the results, initially given 20\% of training data with imbalanced classes, BSL was biased with a TPR of only 29.0\%. Although the TPR of BSS-GAN-0 dropped to 60.7\% compared to using fully labeled dataset, it was still far less biased than BSL. BSL cannot improve beyond this point, as it can only learn from labeled data. However, as we introduced more unlabeled data to BSS-GAN (Figures \ref{fig_Unsup_BSSGAN_40} \& \ref{fig_Unsup_BSSGAN_80}), the TPR of BSS-GAN improved by 4\% and 11\% by supplementing 50\% and 100\% of the unlabeled data respectively. Although the supplementary data do not provide label information, under the semi-supervised learning setting, BSS-GAN can still utilize information from the unlabeled samples. During balanced batch sampling, the number of unlabeled data fed to each batch stays consistent with that of a single-class sub-batch, i.e., $n_{ul}=n_{l}/K$. As a result, even as more unlabeled data are introduced, they do not overwhelm the labeled samples during batch-by-batch training. Thus, unlabeled data, once handled appropriately in each training batch, are able to supplement the learnt features and improve the classifier's performance. 

One seeming caveat is the decreasing TNR of BSS-GAN as more unlabeled data are supplemented. However, the upward trend of the $F_5$ score suggests that BSS-GAN models trained with more unlabeled data are better, which is based on our interest where high recall is prioritized over precision. This experiment shows once more that the overall accuracy is a deceptive measure as the BSL achieved a 95.1\% accuracy compared with BSS-GAN-100's 91.6\% (with 8,160 unlabeled samples), yet the BSL's performance is the worst overall in terms of TPR and $F_5$ score.

\begin{table}
\centering
\caption{\enspace Classification performance in the study of unlabeled data utilization ($\%$)}
\label{tab_Unsup}
\vspace*{-1mm}
\begin{adjustbox}{width=\linewidth}
\begin{tabular}{c|c|c|c|c|c}
\hline
\textbf{Pipeline} & \textbf{Unlabeled data} & \textbf{TPR} & \textbf{TNR} & $\mathbf{F_5}$ & \textbf{Accuracy} \\ \hline
BSL               &  -             & 29.0   & 99.2   & 29.7  & 95.1      \\ \hline
BSS-GAN-0           &  -             & 60.7   & 95.0   & 59.7  & 93.0    \\ \hline
BSS-GAN-50           &  4,080         & 65.0   & 93.2   & 63.2  & 91.5     \\ \hline
BSS-GAN-100           &  8,160         & 72.3   & 92.8   & 70.0  & 91.6      \\ \hline
\end{tabular}
\end{adjustbox}
\end{table}

\subsection{Experiment 5: Comparisons with weighted loss methods}
\begin{figure}
\center
\begin{subfigure}{.4\linewidth}
  \centering
  \includegraphics[width=\linewidth]{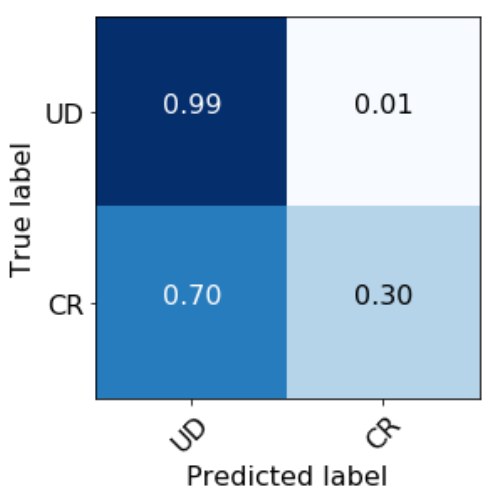}
  \caption{\label{fig_BSL_Rev_Freq_CR16}Crack detection}
\end{subfigure}
\begin{subfigure}{.4\linewidth}
  \centering
  \includegraphics[width=\linewidth]{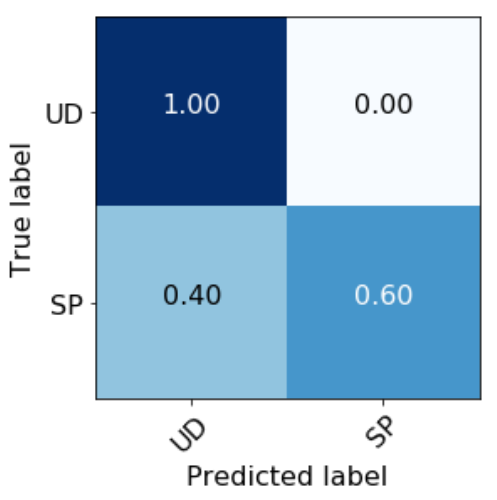}
  \caption{\label{fig_BSL_Rev_Freq_SP32}Spalling detection}
\end{subfigure}%

\begin{subfigure}{.45\linewidth}
  \centering
  \includegraphics[width=\linewidth]{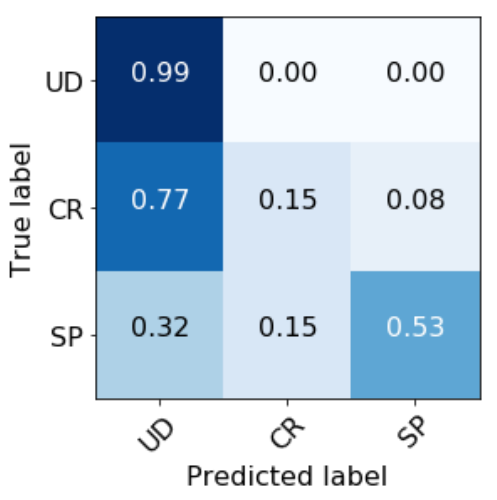}
  \caption{\label{fig_BSL_Focal_TRI}Damage pattern recognition}
\end{subfigure}

\caption{\label{fig_weighted_loss_CM}\enspace Normalized CM in three tasks using BSL with weighted loss methods (BSL-w)}
\end{figure}

In this section, besides the above TL/DA methods, the performance of the BSL using weighted loss function (BSL-w) is investigated and then compared with our BSS-GAN. The best CM results using the weighted loss approach are shown in Figure \ref{fig_weighted_loss_CM}, where Figures \ref{fig_BSL_Rev_Freq_CR16} \& \ref{fig_BSL_Rev_Freq_SP32} and Figure \ref{fig_BSL_Focal_TRI} are from BSL using balanced cross entropy and focal loss, respectively.

Compared to BSL's performance in Figures \ref{fig_CR16_BSL} \& \ref{fig_SP32_BSL}, the one using weighted loss even obtained lower TPR values. From Figures \ref{fig_TRI_BSL} \& \ref{fig_BSL_Focal_TRI}, the model experienced a performance trade-off between CR and SP, where the weighted loss function improved SP's recall from 32\% (BSL) to 53\% (BSL-w) with a decrease of CR's recall from 29\% (BSL) to 15\% (BSL-w). Such performance is no better than BSL, let alone BSS-GAN (Figures \ref{fig_CR16_BSSGAN}, \ref{fig_SP32_BSSGAN} \& \ref{fig_TRI_BSSGAN}). In addition, in the two binary cases, the $F_2$ scores of BSL-w are 34.0\% \& 64.8\%, which are even lower than those of BSL (35.2 \& 66.1\%). Similarly, the $F_5$ scores are 30.7\% \& 60.9\% for BSL-w and lower than those of BSL as well (31.7\% \& 62.8\%).

In summary, similar to the performance of conventional methods, the weighted loss function methods do not improve the classification accuracy. It is inferred from experiment 5 that only manipulating the loss, which intends to lessen the contribution from the majority and emphasize the minority, does not indeed address the class-imbalance issue. On the contrary, our BSS-GAN strictly enforces the class ratio in each training batch, which is shown to be straightforward and efficient.

\subsection{Summary}
Through experiments, the following key conclusions are drawn from the study:
\begin{itemize}
\item[$\checkmark$] In general, BSS-GAN outperformed others in both binary crack and spalling detection under low-data and imbalanced-class settings. It achieved a significant improvement in TPR by reducing FN with only a slight decrease of TNR. BSS-GAN achieved better $F_\beta$, e.g., $F_5$ scores with more weights on recall (TPR) over precision. 
\item[$\checkmark$] Over-sampling the minority class by GAN-generated images (BOS-GAN) led to worse performance than the baseline (BSL) in spalling detection tasks. This is caused by: (1) introduction of extra parameters, (2) subjective manual synthetic image selection, and (3) generation of ``adversarial'' images. These factors caused unstable training behavior and exacerbated BOS-GAN's bias in favor of the majority class (UD). Such observations correlate to the findings in \citep{gao2019deep}.
\item[$\checkmark$] BUS, BOS-DA \& BSL-SDF had similar but limited improvements over BSL, which are not satisfactory in practice. Their flaws include: in BUS, under-sampling eliminated a large portion of the labeled majority-class data, causing information loss; in BOS-DA, the conventional DA failed to increase feature variety; in BSL-SDF, the model did not sufficiently address the imbalanced-class issue although it improved parameter initialization.
\item[$\checkmark$] In three-class classification, all pipelines except BSS-GAN were prone to predicting CR as UD (leading to low CR recalls). On the contrary, BSS-GAN obtained a promising CR recall of about 70\%, while maintaining a high SP recall of 94\% and a good UD recall of 91\%. This further indicates the stable and high potential of the BSS-GAN for handling imbalanced multi-class tasks.
\item[$\checkmark$] BSS-GAN generated images of all classes without mode collapse, because it learned from a mixed-class distribution with balanced batch sampling. If only concerning generated image quality by human visual judgement, the ordinary GAN generator used in BOS-GAN and BSL-SDF was slightly better. It is inferred that the improvement of BSS-GAN's discriminator weakened its generator, but the generator was yet able to generate realistic images for the classifier to learn new features from.
\item[$\checkmark$] When labeled data have limited availability, the semi-supervised setting of the BSS-GAN allows it to utilize unlabeled data. With a proper ratio of unlabeled data placed into each training batch, BSS-GAN was able to capture meaningful information from the unlabeled data.
\item[$\checkmark$] It is found that as an alternative method to address class-imbalance issues, applying weighted loss function methods to the BSL did not improve the accuracy in our experiments, which further indicates the robustness and effectiveness of the proposed BSS-GAN under low-data imbalanced-class conditions.
\end{itemize}

However, several aspects need to be investigated further:
\begin{itemize}
\item[$\circ$] Even though pursuing a high precision is not so meaningful in extreme imbalanced-class problems, reducing FP is still desired, in order to lower the costs of false alarms. From the experimental results, there exists a room for FP reduction from BSS-GAN.
\item[$\circ$] Our experiments have only shown that BSS-GAN is effective under the low-data regime. More experiments need to be conducted with medium-data and big-data regimes using other open-source datasets than the used PEER Hub ImageNet ($\phi$-Net), including those outside the SHM domain, e.g., road damage \citep{maeda2018road}, to explore the range of effective applications of the BSS-GAN.
\item[$\circ$] Only one general-purpose CNN architecture was tested in our experiments. Parametric studies with respect to the network architectures and comparative experiments with other GAN-based methods, e.g., BAGAN \citep{mariani2018bagan}, are recommended in future studies.
\item[$\circ$] Besides classification, damage localization and segmentation also face the issues of data deficiency and class imbalance. This suggests other uses of the BSS-GAN.
\item[$\circ$] Other powerful classification algorithms modified under the BSS-GAN framework, e.g., neural dynamic classification \citep{rafiei2017new}, dynamic ensemble learning \citep{alam2020dynamic}, and finite element machine for fast learning \citep{pereira2020fema}, will be considered in future studies. 
\end{itemize}

\section{Conclusions}
In this study, we discussed three key issues that impede the real-world applications of DL in vision-based structural damage assessment, namely data deficiency, class imbalance, and limited computing power. To address these issues, BSS-GAN, a semi-supervised learning GAN pipeline with balanced batch sampling, was proposed. It is an alternative to conventional or GAN-based data augmentation methods. To verify the effectiveness and efficiency of BSS-GAN in classification tasks, a series of computer experiments related to crack and spalling detection of reinforced concrete were designed and conducted under low-data and imbalanced-class regimes. In addition, computing power limitations were simulated via using a shallow and generic CNN as the base design for all pipelines. The experimental results were analyzed and compared with other five pipelines based on multiple metrics, i.e., recall (TPR) and $F_\beta$ scores ($F_2$ \& $F_5$). The synthetic image generation capabilities were then compared between BSS-GAN and the ordinary GAN. Lastly, the effectiveness of supplementing unlabeled data for feature learning in BSS-GAN was investigated. The results demonstrate that the BSS-GAN is able to achieve better performance detection over conventional pipelines in all designed experiments. In conclusion, the promising results shed light on the high potential of semi-supervised GAN in vision-based damage assessment and SHM. This is clearly worth significant research efforts in the future.

\section*{Acknowledgements}
The authors acknowledge the funding from Tsinghua-Berkeley Shenzhen Institute (TBSI), Taisei Chair of Civil Engineering, Berkeley Education Alliance for Research in Singapore (BEARS) for the Singapore Berkeley Building Efficiency and Sustainability in the Tropics (SinBerBEST) program, and USDA AI Institute for Next Generation Food Systems (AIFS) (USDA award number 2020-67021-32855).

\nomenclature{BSL}{Baseline}
\nomenclature{DA}{Data Augmentation}
\nomenclature{UD}{Undamaged damage state}
\nomenclature{CR}{Cracked}
\nomenclature{SP}{Spalling}
\nomenclature{SDF}{Synthetic Data Fine-tuning}
\nomenclature{CM}{Confusion Matrix}
\nomenclature{TP(R)}{True Positive (Rate)}
\nomenclature{TN(R)}{True Negative (Rate)}
\nomenclature{FP}{False Positive}
\nomenclature{FN}{False Negative}


\printnomenclature

\balance
\nocite{*}
\bibliography{wileyNJD-APA}%

\end{document}